\documentclass[letterpaper]{article} 
\usepackage{aaai25}  
\usepackage{times}  
\usepackage{helvet}  
\usepackage{courier}  
\usepackage[hyphens]{url}  
\usepackage{graphicx} 
\usepackage{natbib}  
\usepackage{caption} 
\usepackage{algorithm}
\usepackage{algorithmic}
\usepackage{amsmath}
\usepackage{amssymb}
\usepackage{booktabs}
\usepackage{makecell}
\usepackage{multirow}
\usepackage{graphicx}
\usepackage{adjustbox}
\usepackage{amssymb}
\usepackage{color}
\usepackage{nameref}
\usepackage{subcaption}
\usepackage{newfloat}
\usepackage{listings}
\DeclareCaptionStyle{ruled}{labelfont=normalfont,labelsep=colon,strut=off} 
\floatstyle{ruled}
\newfloat{listing}{tb}{lst}{}
\floatname{listing}{Listing}
\title{Accurate and Regret-aware Numerical Problem Solver for Tabular Question Answering}
\author{
    Yuxiang Wang,
    Jianzhong Qi\footnote{Corresponding author},
    Junhao Gan
}
\affiliations{
    School of Computing and Information Systems, The University of Melbourne\\

    yuxiang.wang8@student.unimelb.edu.au, \{jianzhong.qi, junhao.gan\}@unimelb.edu.au
}

\usepackage{bibentry}

\begin{document}

\maketitle

\newcommand{\model}{TabLaP}
\newcommand{\nmodel}{NumSolver}
\newcommand{\cmodel}{AnsSelector}
\newcommand{\vmodel}{TwEvaluator}
\newcommand{\datasetname}{FTQ}
\newcommand{\wikidataset}{WTQ}

\begin{abstract}
Question answering on free-form tables (a.k.a. \emph{TableQA}) is a challenging task because of the flexible structure and complex schema of tables. Recent studies use Large Language Models (LLMs) for this task, exploiting their capability in understanding the questions and tabular data, which are typically given in natural language and contain many textual fields, respectively. While this approach has shown promising results, it overlooks the challenges brought by numerical values which are common in tabular data, and LLMs are known to struggle with such values.  We aim to address this issue, and we propose a model named \model\  that uses \underline{L}LMs \underline{a}s a \underline{p}lanner rather than an answer generator. This approach exploits LLMs' capability in multi-step reasoning while leaving the actual numerical calculations to a Python interpreter for accurate calculation. Recognizing the inaccurate nature of LLMs, we further make a first attempt to quantify the trustworthiness of the answers produced by \model, such that users can use \model\ in a \emph{regret-aware} manner. Experimental results on two benchmark datasets show that \model\ is substantially more accurate than the state-of-the-art models, improving the answer accuracy by 5.7\% and 5.8\% on the two datasets, respectively. 
\end{abstract}

%

\section{Introduction}

\begin{figure}[ht!]
    \centering
    \begin{subfigure}[b]{\linewidth}
        \centering
        \includegraphics[width=\linewidth]{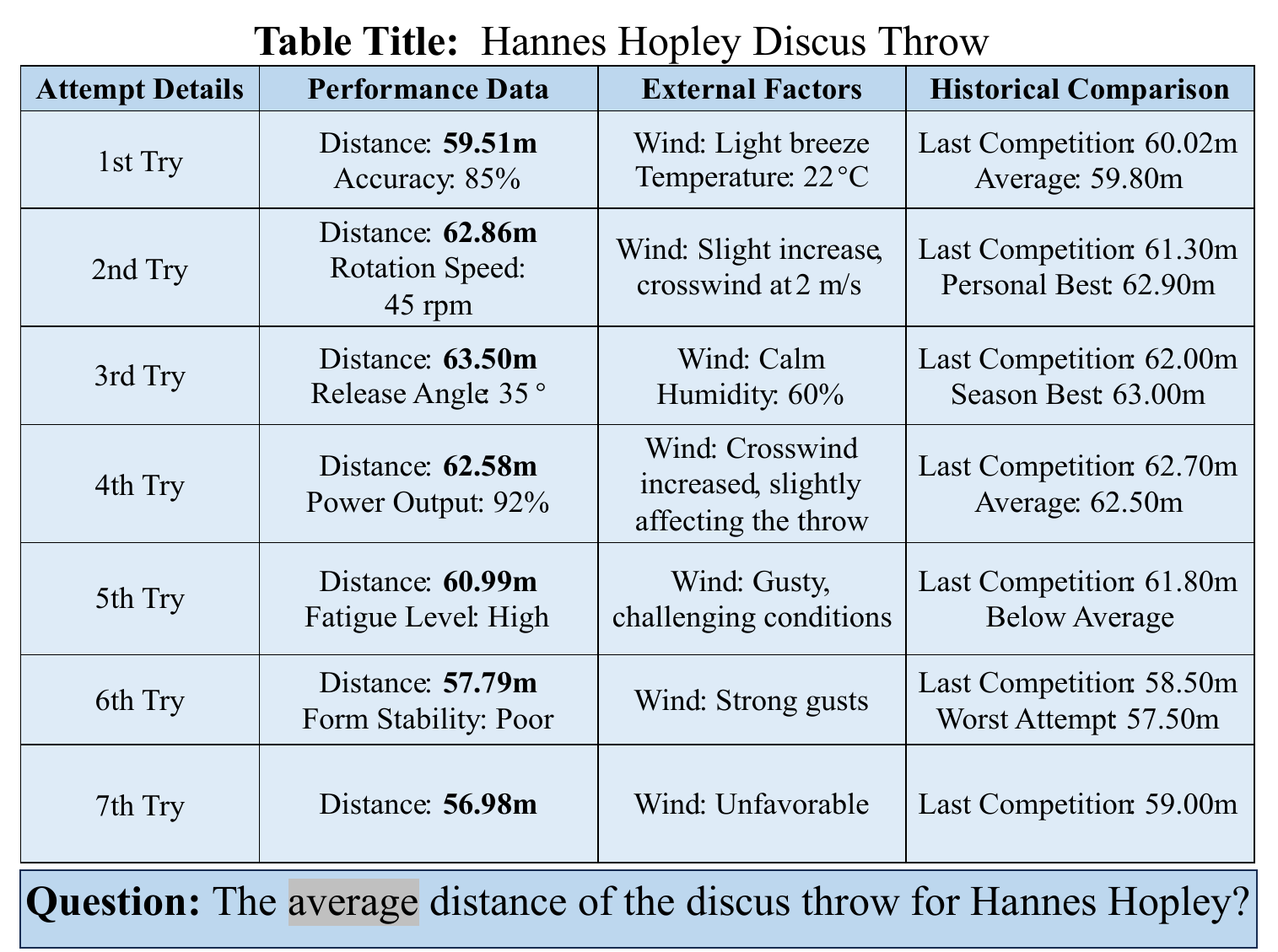}
       \caption{}
        \label{fig:inputs}
    \end{subfigure}
    \begin{subfigure}[b]{\linewidth}
        \centering
        \includegraphics[width=\linewidth]{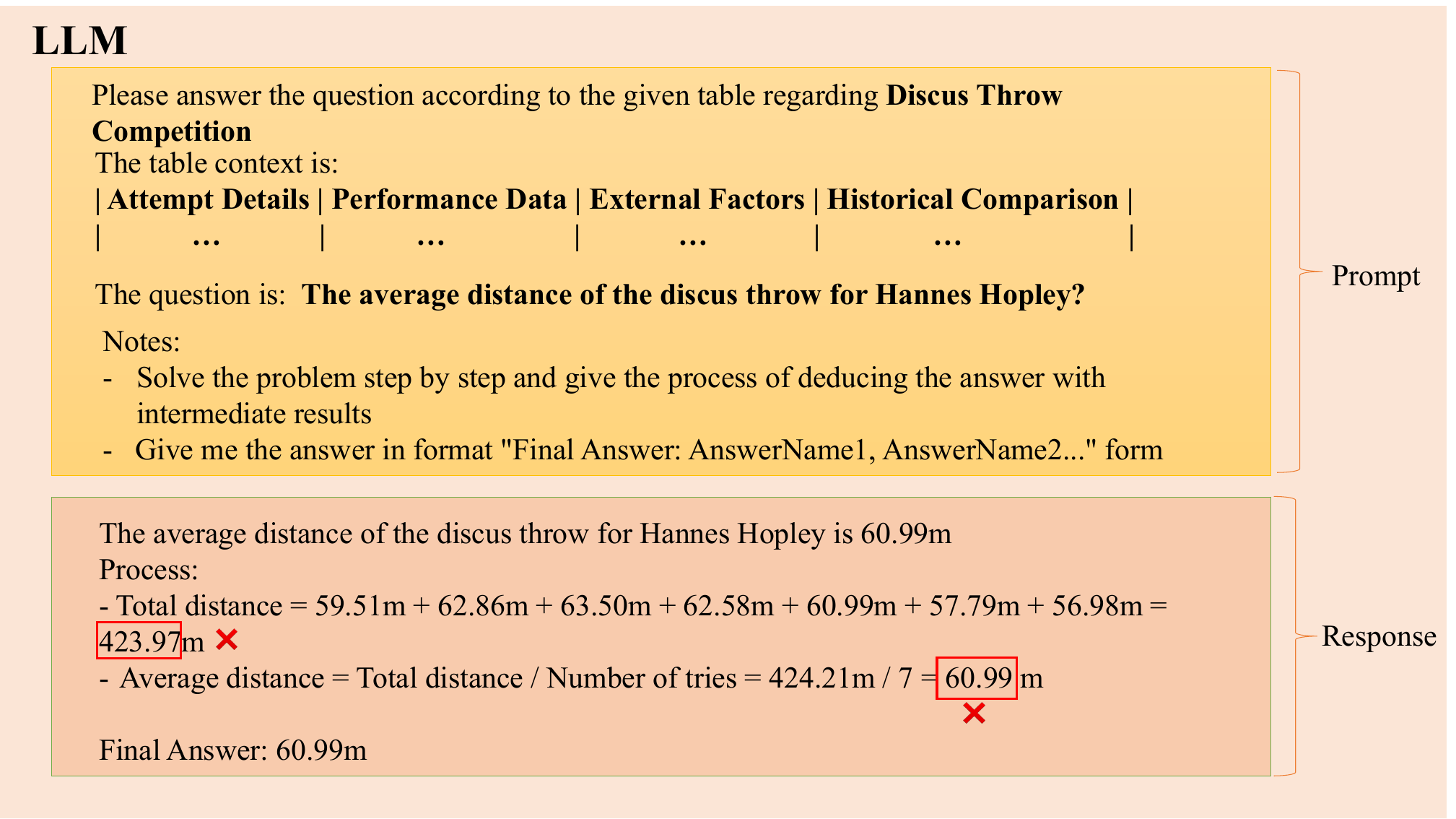}
       \caption{}
        \label{fig:direct_prompt}
    \end{subfigure}
    \begin{subfigure}[b]{\linewidth}
        \centering
        \includegraphics[width=\linewidth]{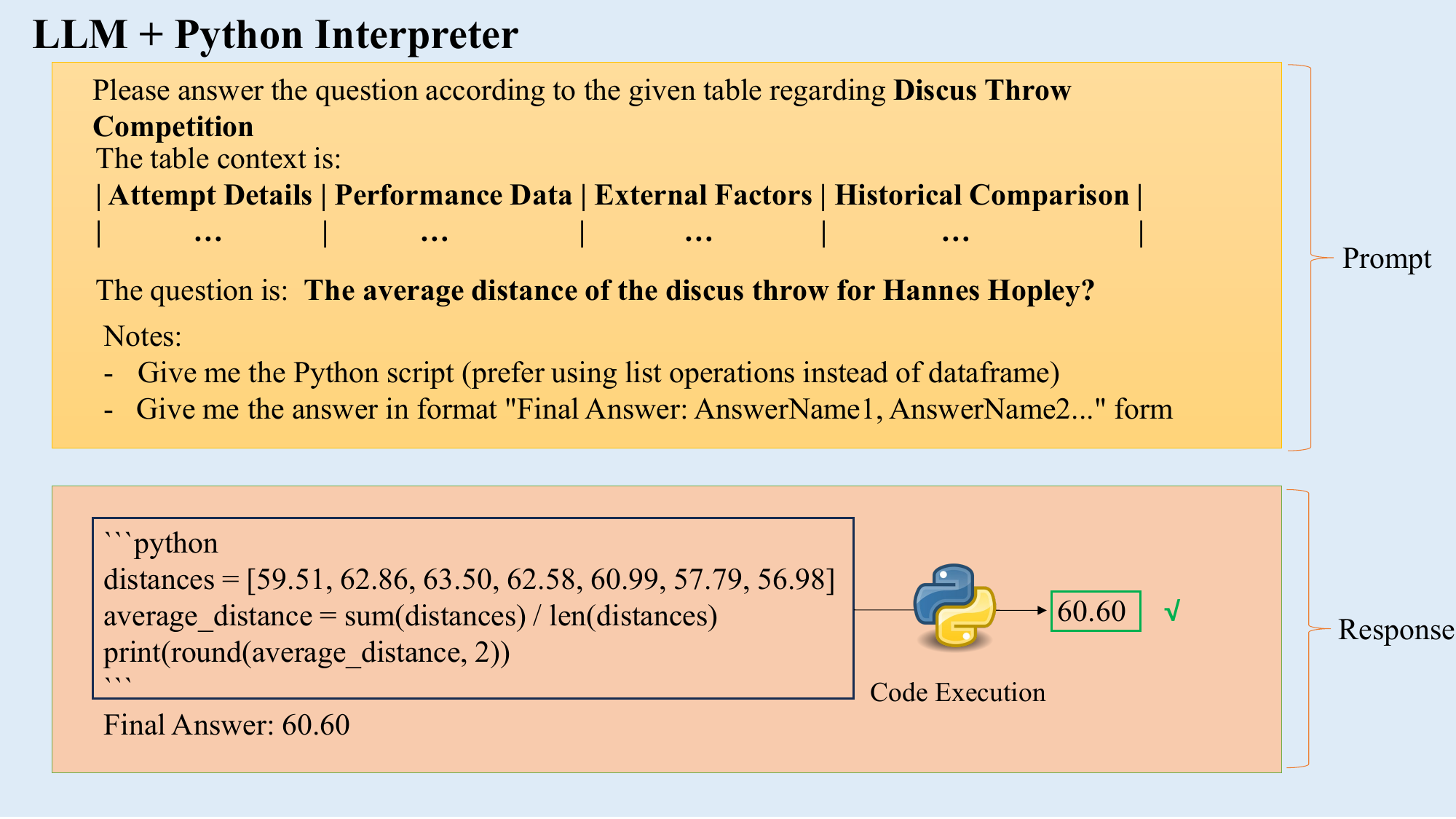}
      \caption{}
        \label{fig:llm_python}
    \end{subfigure}
    \caption{A TableQA example with a numerical question: (a)~Input table and question; (b)~Answer from prompting an LLM (GPT-3.5 Turbo) directly; (c)~Answer from prompting the LLM to generate a Python script for the calculation (we prompt the LLM to ignore the unit to suit the evaluation procedure of the benchmark datasets, which can be kept by adapting the prompt easily).}
    \label{fig:numerical_example}
\end{figure}

Tables are a commonly used data format to organize and present information. \emph{Table Question Answering} (TableQA) aims to answer questions based on data in tables. It arises as an important problem to automatically extract information from free-form tables for non-experts \cite{ye2303comprehensive}. In a typical TableQA task, questions are given in natural language. The input tables are in free-form and may not have a well-defined schema, i.e., they are not necessarily relational tables, such as web tables \cite{cafarella2008webtables}.
In other words, these tables may have mixed data types in columns and non-predefined logical forms \cite{jin2022survey}. In this case, structural query language-based solutions~\cite{pasupat-liang-2015-compositional, zhong2017seq2sql, pourreza2023dinsqldecomposedincontextlearning, gao2023texttosqlempoweredlargelanguage} are not always applicable. See Figure~\ref{fig:inputs} for an example of the TableQA task. 

Recent studies~\cite{cheng2023binding, ye2023large, liu-etal-2024-rethinking} use Large Language Models (LLMs), exploiting their semantic  capabilities to analyze textual questions as well as tabular data,  where many fields are often in text. 
Unfortunately, these studies have overlooked a notorious issue of the LLMs -- their limited capability in handling numerical data \cite{frieder2024mathematical}. 
As Figure~\ref{fig:direct_prompt} shows, prompting an LLM (GPT-3.5 Turbo) directly with a table and a question for numerical calculation could lead to an inaccurate answer.

To address this issue, we study how to strengthen the capability of LLMs to handle numerical questions. 
A crucial observation behind our proposed solution is that while LLMs struggle with carrying out numerical calculations, they are capable of producing plans to execute such calculations. 
As Figure~\ref{fig:llm_python} shows, using the same LLM but prompting it to generate a calculation plan in Python, the generated Python script can be executed and yield the correct answer. 

Based on this observation, we propose a \underline{Tab}leQA model with an \underline{L}LM \underline{a}s a \underline{p}lanner, \emph{\model}\footnote{Source code available at \url{https://github.com/yxw-11/TabLaP}}.  We call the planner LLM \emph{\nmodel} and design a prompt to guide it to generate a plan that decomposes a complex  numerical question into sequential steps (in Python script) based on Chain-Of-Thought~\cite{wei2022chain}. The script is then executed by a Python interpreter to produce an answer to the question.

Moreover, to largely retain the strong capability of LLMs to process non-numerical questions, \model\ takes a dual-branch structure, where \nmodel\ forms a branch and a state-of-the-art (SOTA) TableQA model forms the other.  

As each of the two branches produces an answer, to integrate the answers from them, we exploit a third LLM (named \emph{\cmodel}) -- an open-source one (LlaMa 3-{8B-}Instruct) which allows for fine-tuning  --  to take the question and answers from both model branches (including the reasoning texts) as input and selects a branch to trust. The answer from the selected branch is then returned as the final answer. 

Recognizing the inaccurate nature of LLMs, we further quantify the \emph{trustworthiness} of the answers generated by \model. We propose a module named \emph{\vmodel} based on yet another LLM and Multi-Arm Bandit (MAB) \cite{vermorel2005multi}. \vmodel\ tracks the answer correctness of both \model\  branches over time and yields a trustworthiness label of the final answer accordingly. This label enables users to consume the answer in a \emph{regret-aware} manner. 

Overall, this paper makes the following contributions:

\begin{itemize}
\item We propose \model, a multi-LLM-based model for TableQA tasks with both numerical and non-numerical questions. The core contribution of \model\ lies in its holistic system design that exploits the strength of each model forming a module of \model, rather than forcing a model onto an unsuitable task.

\item We propose \nmodel, an LLM-based module to process numerical questions. 
We further fine-tune an open-source LLM-based \cmodel\ module to decide whether to use the answer generated by \nmodel\ or a SOTA TableQA model (e.g.,~\citealp{liu-etal-2024-rethinking}),  exploiting their capabilities in answering numerical and non-numerical questions, respectively.

\item We propose a regret-aware  scheme enabled by an LLM-and-MAB-based module named \vmodel\ that tracks the correctness of the answer generation modules and produces a trustworthiness label for the answers.

\item We conduct experiments to test the effectiveness of \model\ on \texttt{WikiTableQuesetions}~\cite{pasupat-liang-2015-compositional} and \texttt{\datasetname}. 
 \texttt{WikiTableQuesetions} is a public dataset, while \texttt{\datasetname} is adapted by us from the \texttt{FeTaQA} dataset~\cite{nan2022fetaqa} by removing answer tokens non-directly relevant to the questions.
The experimental results show that \model\ outperforms SOTA TableQA models on both datasets by 5.7\% and 5.8\% in accuracy, respectively. Meanwhile, the answer trustworthiness labels
generated by \model\ help reduce the user regret ratio on
using the model generated answers by 
19.6\% and 20.6\%
on the two datasets, respectively, compared with always
trusting the model generated answers.
\end{itemize}

\section{Related Work}\label{sec:related-work}
Studies on TableQA are mainly driven by designing models that can understand questions in natural language and tabular data. 
These studies can be categorized into \emph{semantic parsing-based}, \emph{pre-trained language model (PLM)-based}, and \emph{large language model (LLM)-based}.

\textbf{Semantic parsing-based methods.}  Semantic-parsing-based methods transform natural language questions into a logical form (e.g., SQL) that machines can understand  and execute. There are two sub-categories of methods: (i)~\emph{weakly-supervised}~\cite{pasupat-liang-2015-compositional,neelakantan2017learning,yu2018typesql},  and (ii) ~\emph{fully-supervised}, such as NL-to-SQL~\cite{zhong2017seq2sql,pourreza2023dinsqldecomposedincontextlearning,gao2023texttosqlempoweredlargelanguage}. For weakly supervised methods, a semantic parser generates the logical form based on an input question, a table, and the answer. There is no pre-defined ground-truth logical form. Fully supervised methods, on the other hand, further take a ground-truth logical form as their input. Both sub-categories focus on analyzing the questions. They are less effective on tables with complex structures and data types \cite{hong2024nextgenerationdatabaseinterfacessurvey}. 

\textbf{PLM-based methods.} Language model-based methods, including PLM-based ones, focus on guiding the models to understand tabular data. There are two directions: (i)~Tailoring Transformer~\cite{vaswani2017attention} model structure for better tabular data understanding. For example, TAPAS~\cite{herzig-etal-2020-tapas} masks table cells and extends BERT~\cite{devlin-etal-2019-bert}  by adding column, row, and rank embeddings. TUTA~\cite{wang2021tuta} further masks columns or table segments and designs a special attention mechanism with a tree structure to capture the hierarchical relationships and dependencies for tabular data. TaBERT~\cite{yin-etal-2020-tabert} extends BERT with \emph{vertical self-attention} and combines text and table representations in a unified framework. (ii)~Pre-training and fine-tuning models for end-to-end TableQA. For example, TAPEX~\cite{liu2022tapex} pre-trains BART \cite{lewis-etal-2020-bart} with a large synthetic dataset derived from the \texttt{WikiTableQuestions} dataset. CABINET~\cite{patnaik2024cabinet} fine-tunes content relevance-based modules that reduce the weight of irrelevant rows or columns to guide the PLM focus more on the key information. OmniTab \cite{jiang-etal-2022-omnitab} also leverages BART as its backbone model, while it is pre-trained using both real and synthetic data -- these include SQL queries from the Spider dataset ~\cite{yu-etal-2018-spider} and synthetic natural language sentences converted from the SQL queries using their proposed SQL2NL model. The aim is to enhance the model capability to perform few-shot learning for TableQA. The PLM models are smaller than the latest LLMs. They have limited capabilities in understanding tabular semantics, leading to sub-optimal TableQA results, as shown  empirically.

\textbf{LLM-based methods.} 
Recent studies exploit the semantic and context (i.e., table) tracking capabilities of LLMs. 

In-context learning is a common method, where a few TableQA examples are fed into an LLM together with an input question and a table to prompt the LLM for answer generation. For example,  Chain-of-Table~\cite{wang2024chainoftable} solves TableQA problems step by step and obtains a sub-table at each step using a GPT-based model. DATER~\cite{ye2023large} decomposes tables and questions into sub-tables and sub-questions at each step to help LLMs understand the question and data more easily.

Mix-SC \cite{liu-etal-2024-rethinking}, the SOTA model, generates a few answers for each question exploiting the stochastic nature of LLMs. It uses  two types of prompts: prompting the LLM to run as a Python agent to execute Python scripts directly and Chain-of-Thoughts prompting to ask the LLM to solve problems stey by step~\cite{wei2022chain}. The best answer is returned, which is selected using a \emph{self-consistency} method, i.e., taking the most frequent answer. The models above generate answers by LLMs directly, while we use LLMs to generate an answer calculation plan. These models can form a branch in our model. We use Mix-SC for its SOTA performance. 

Binder~\cite{cheng2023binding} uses an LLM to generate an initial program for an input question and identify portions of the program that are difficult to solve. It then re-invokes the LLM to supplement these parts. Finally, the refined program is executed by an interpreter to obtain the TableQA answer.
 LEVER~\cite{ni2023lever} exploits an LLM as a planner to generate multiple possible SQL queries, executes them to obtain candidate answers, and  employs a verification model to select the best answer. \model\ differs from Binder in that it uses the Chain-of-Thought method, allowing the LLM to generate problem-solving scripts step by step and enables the LLM to directly analyze and produce an answer, mitigating the impact of code execution errors. In comparison, Binder's answers heavily rely on the quality of the initially generated program, while its design does not account for handling errors in the program. Therefore, \model\ better leverages the analytical capabilities of LLMs and provides a more reliable solution. Compared with LEVER, \model\ generates Python scripts instead of SQL queries, which are more flexible for non-relational tables. Further, \model's answer selector examines not only  the answers (as done by LEVER) but also the reasoning process of the answers, leading to more accurate answer selection. 

\textbf{Numerical Reasoning for LLMs.} Understanding numerical data and performing calculations are known issues with LLMs~\cite{didolkar2024metacognitive, frieder2024mathematical}. 

\begin{figure*}[t!]
     \centering
     \includegraphics[width = 1\linewidth]{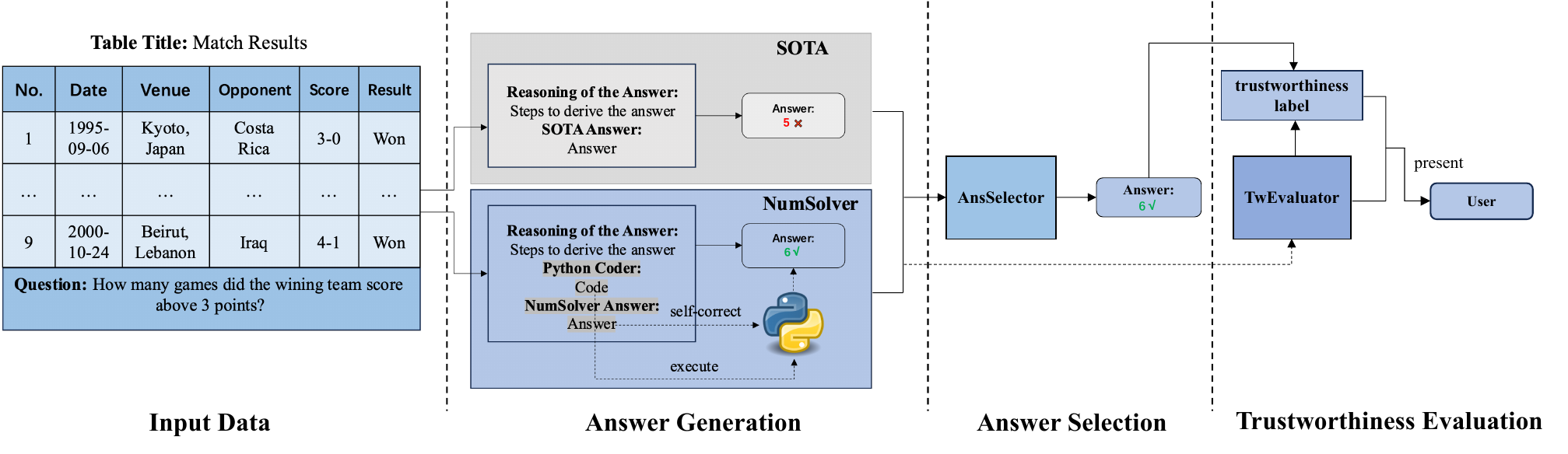}
     \caption{Overview of \model. The model has three stages: answer generation, answer selection, and trustworthiness evaluation. (i)~The answer generation stage uses both a SOTA TableQA model~\cite{liu-etal-2024-rethinking} and our \nmodel\ module to generate answers for an input question and a table, where \nmodel\ focuses on numerical questions. (ii)~The answer selection stage selects answers from the  generated ones, based on the question and the reasoning steps generated by the two models. (iii)~The trustworthiness evaluation stage tracks the success rates of both the SOTA model and \nmodel, and generates a trustworthiness label that is presented to users together with the selected answer.}
     \label{fig:tabmi}
\end{figure*}

Efforts have been made to improve LLMs' mathematical capability. DELI~\cite{zhang2023evaluating}, which utilizes an LLM to generate the solution process for mathematical problems, identifies mathematical expressions within the solution, and then calls external tools to execute the calculation. PAL~\cite{gao2023pal} breaks down a mathematical problem into multiple steps, generates executable code for each step to produce intermediate results, and combines these intermediate answers to calculate the final answer. Inspired by this latter study, we generate multiple answers for an input question and train a classifier to select the best answer. Our solution differs from both studies above, in that we use an LLM to generate Python scripts and execute them end-to-end, effectively reducing the impact of errors produced by the intermediate steps. Additionally, by analyzing the reasoning process to select the best answer, \model\ addresses scenarios where the calculation is correct while the reasoning is flawed.

\section{Methods}
Given a table $T$ and a question $Q$ regarding the data in $T$, our goal is to design a model that produces an accurate answer for $Q$. Here, $Q = (q_1, q_2, \ldots, q_{|Q|})$ is given in natural language, where $q_i \in Q$ ($i \in [1, |Q|]$) is a token (e.g., a word), $T$ is  also represented as a sequence of tokens in natural language, where the  cells are separated  by special characters such as `$\vert$' while the rows are separated by newline characters. The effectiveness of a model is measured by a token-based comparison between the answer generated by the model and the ground truth. 

In this paper, we are particularly interested in multi-step numerical questions which require applications of two or more basic arithmetic and relational operators \{$+$, $-$, $*$, $/$, $>$, $<$ \}. These operators are common in our experimental datasets, while the techniques proposed can be extended to support more  relational operators such as  $\ge$ and $\le$ straightforwardly.
It is known that existing LLMs are less effective in handling this type of questions. 


\subsection{Model Overview} 

As outlined in Algorithm~\ref{alg:tabmi}, 
there are three stages in the question answering process of our \model\ model: (i) \emph{answer generation}, (ii) \emph{answer selection}, and (iii) \emph{answer trustworthiness evaluation}. 

In the answer generation stage, two models are adopted. One is a SOTA model~\cite{liu-etal-2024-rethinking} (detailed in~\nameref{sec:related-work}), denoted by $M_{SOTA}$, and the other is our numerical question answering module \nmodel, denoted by $M_{NS}$ (detailed in~\nameref{sec:ans_gen}). These two models take the input pair $(T, Q)$ and individually generate answers (together with their reasoning steps) for $Q$. 

In the answer selection stage, our \cmodel\ module (detailed in~\nameref{sec:ans_select}) takes as input  the question $Q$, the schema of $T$ and the two answers along with the corresponding reasoning steps provided by the two answer generation models in the previous stage. It decides an answer from the two candidates. 

In the answer trustworthiness evaluation stage, our \vmodel~(detailed in~\nameref{sec:ans_verf}) tracks the reliability of the answers generated by the two answer generation models and produces a trustworthiness label for the answer selected in the previous stage. 

\begin{algorithm}[ht!]
\caption{Table Question Answering with \model{}}
\label{alg:tabmi}
\small
\begin{algorithmic}[1]
\REQUIRE Table $T$, questions $Q$, answer generation models $M_{SOTA}$ and \nmodel\ $M_{NS}$, \cmodel,  \vmodel{}
    \STATE \textbf{Answer Generation:} 
    \item[] (1) Generate $\text{prompt}_{SOTA}$ and $\text{prompt}_{NS}$ for $M_{SOTA}$ and $M_{NS}$, respectively 
    \item[] (2) Obtain answer and reasoning pairs $(\text{Ans}_A, \text{Rsn}_A)$ and $(\text{Ans}_B, \text{Rsn}_B)$ by feeding $\text{prompt}_{SOTA}$ and  $\text{prompt}_{NS}$ to $M_{SOTA}$ and $M_{NS}$, respectively
    \STATE \textbf{Selection by \cmodel:} 
   \item[] (1) Combine table $T$ schema and  $Q$ with $(\text{Ans}_A, \text{Rsn}_A)$ and $(\text{Ans}_B, \text{Rsn}_B)$ to form answer selection $\text{prompt}_\text{sel}$ 
   \item[] (2) $\text{Ans}_{\text{best}}$ $\gets$ \cmodel$(\text{prompt}_\text{sel})$ 
    \STATE \textbf{Verification by \vmodel{}:}
   \item[] (1) Combine table $T$  schema and  $Q$ with $(\text{Ans}_A, \text{Rsn}_A)$ and $(\text{Ans}_B, \text{Rsn}_B)$ to form verification $\text{prompt}_\text{twe}$ 
    \item[] (2) $\text{Label}_{\text{true/false}}$ $\gets$ \vmodel{}{}$(\text{prompt}_\text{twe})$ 
    \item[] (3) Update \vmodel\ based on $\text{Ans}_{\text{best}}$, $\text{Label}_{\text{true/false}}$, and ground truth answer

\RETURN $\text{Ans}_{\text{best}}$, $\text{Label}_{\text{true/false}}$
\end{algorithmic}
\end{algorithm} 

\subsection{Answer Generation}\label{sec:ans_gen} 
We use \nmodel~and a SOTA model, Mix-SC~\cite{liu-etal-2024-rethinking}, as two separate branches to answer $Q$.

In \nmodel{}, we prompt a backbone LLM to answer $Q$ step by step and generate the reasoning process with  intermediate results. Besides, we also prompt the LLM to write down the Python scripts to answer $Q$. By executing  the Python scripts with a Python interpreter, question answers are obtained, which are found to be more accurate than those obtained from the direct LLM reasoning process, especially for multi-step numerical questions. When the Python-based answers are different from the LLM-generated ones, and they contain numerical values, priority is given to the Python-based answers. When errors occur during Python execution, the answers obtained directly from the LLM are used. This process is called \emph{self-correctness}.

\begin{figure*}[ht!]
     \centering
     \includegraphics[width = 0.75 \textwidth, height=6cm]{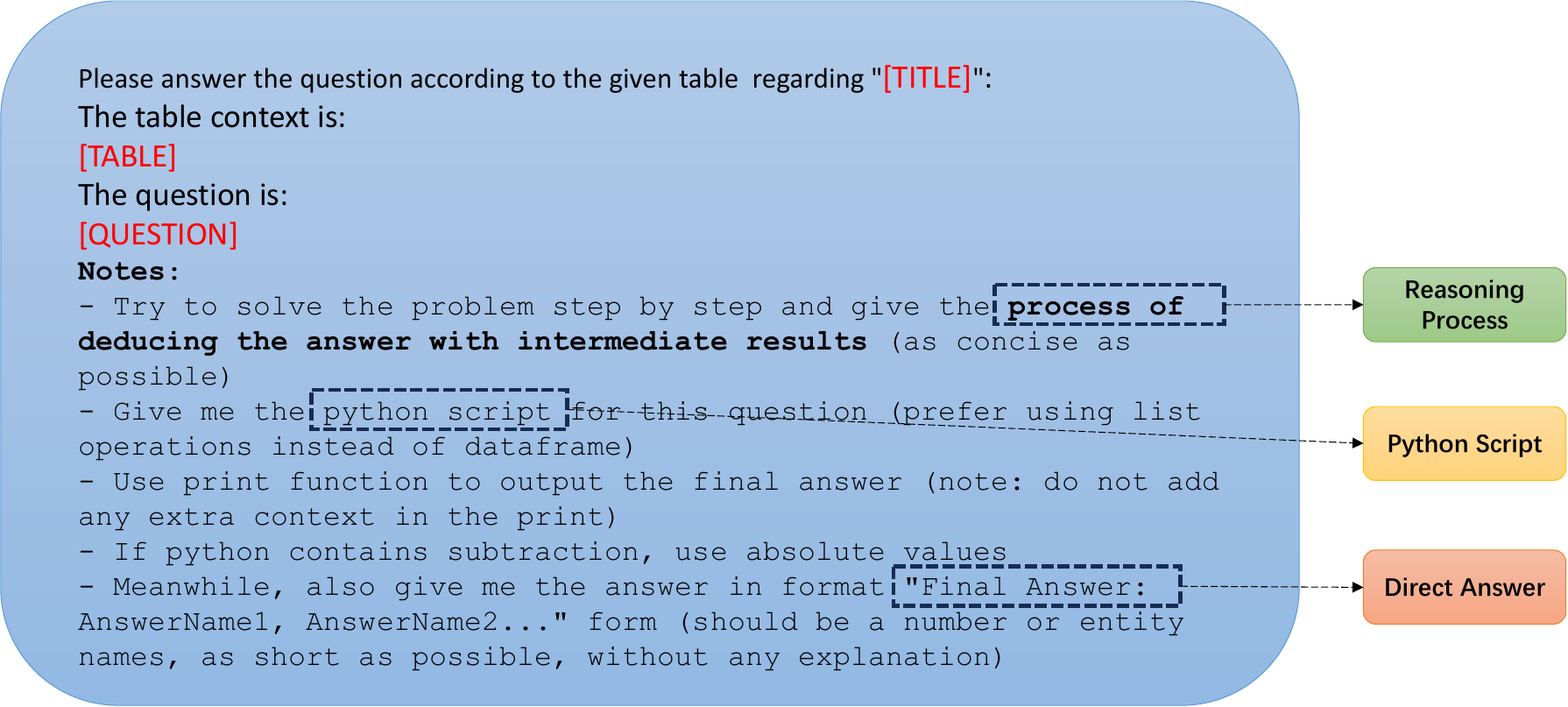}
     \caption{Prompt for the \nmodel{}.}
     \label{fig:math_solver}
\end{figure*}

\begin{figure*}[ht!]
     \centering
     \includegraphics[width = 0.75 \linewidth, height=6cm]{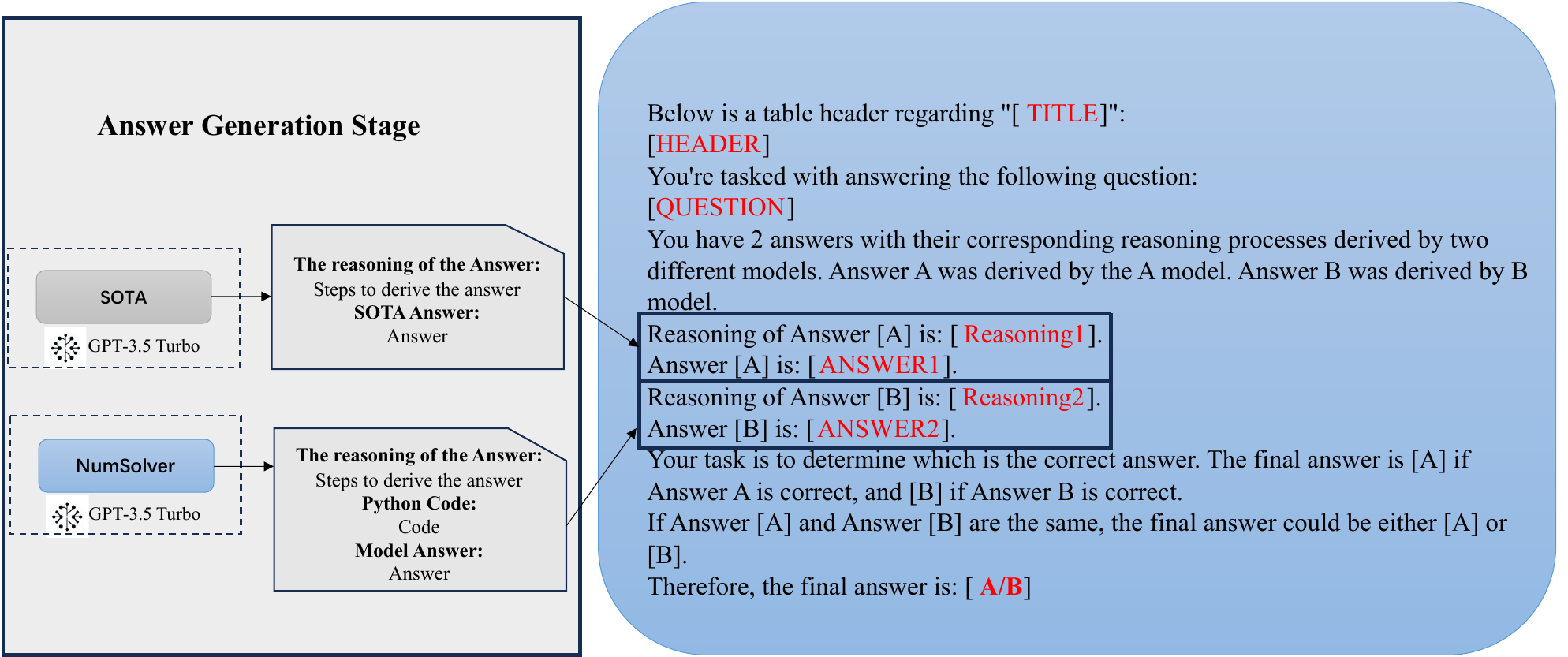}
     \caption{Prompt for the \cmodel.}
     \label{fig:tab_classifier}
\end{figure*}

\begin{figure*}[ht!]
     \centering
     \includegraphics[width = 0.75 \linewidth, height=6cm]{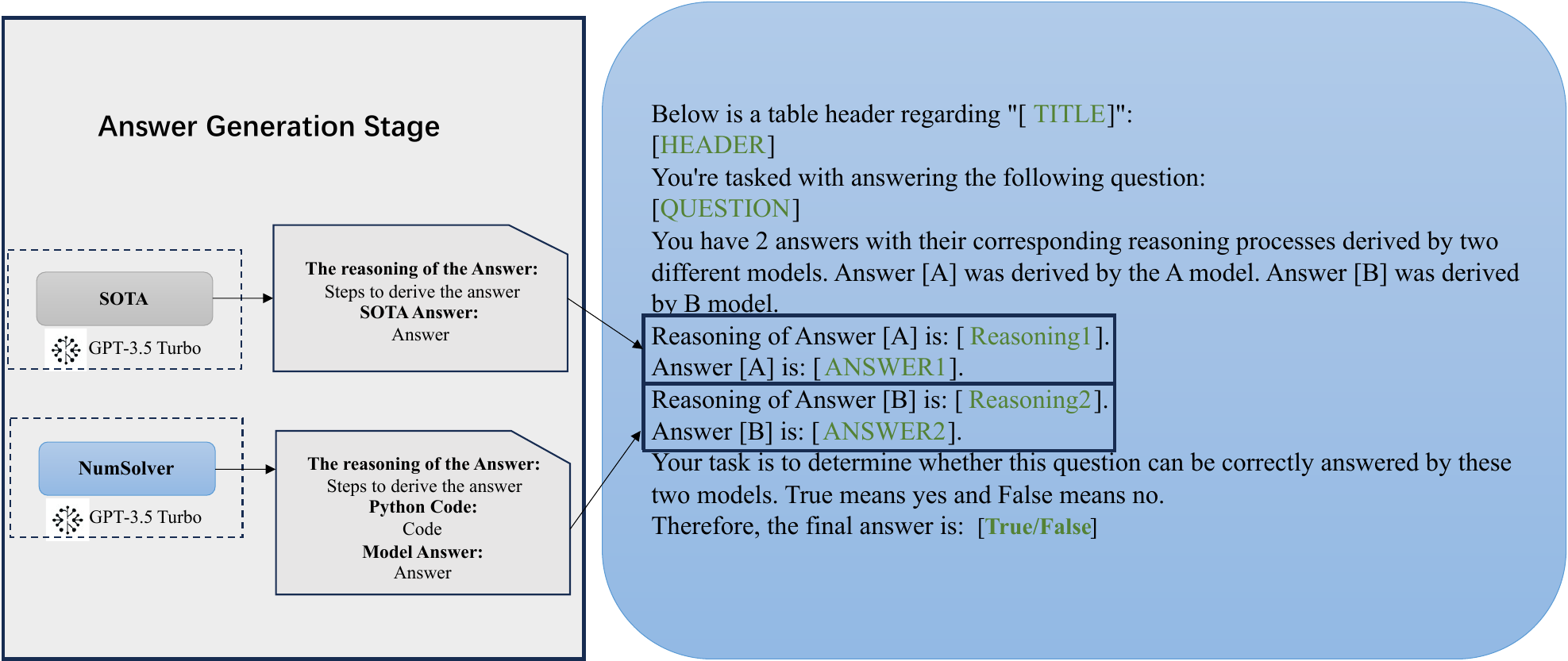}
     \caption{Prompt for the \vmodel{}.}
     \label{fig:tab_verifier}
\end{figure*}

The prompts used in our models (including the ones introduced below) are shown below:

\textbf{Prompt for \nmodel{}.}
For \nmodel, we ask GPT-3.5 Turbo to generate the reasoning process, Python script, and a question answer (obtained with Chain-of-Thought). The prompt template is shown in Figure~\ref{fig:math_solver}.

\textbf{Prompt for \cmodel.} For \cmodel, we prompt a fine-tuned Llama3-8B-Instruct with the reasoning process obtained from the answer generation stage, the answers, and the table information. The prompt template is shown in Figure~\ref{fig:tab_classifier}. \cmodel\ returns a label of either \texttt{[A]} or \texttt{[B]}, indicating that the answer from the SOTA TableQA branch or our \nmodel\ is preferred, respectively.

\textbf{Prompt for \vmodel.} The LLM (i.e., a fine-tuned Llama3-8B-Instruct) of \vmodel\ uses a similar input to that of \cmodel, as shown in Figure~\ref{fig:tab_verifier}. It returns an answer of either \texttt{[True]} or \texttt{[False]}, indicating whether \model\ has answered the input question correctly. 

\subsection{Answer Selection}\label{sec:ans_select}
\cmodel\ decides the best answer from those generated by the two model branches in the answer generation stage, based on the question, table schema, generated answers, and the reasoning process. We use a fine-tuned Llama3-8B-Instruct as \cmodel, which returns a label of either \texttt{[A]} or \texttt{[B]}, indicating that the answer from the SOTA TableQA branch or \nmodel\ is preferred, respectively.

\subsection{Trustworthiness Evaluation}\label{sec:ans_verf}

\vmodel\ trains an LLM (a fine-tuned Llama3-8B-Instruct) with a similar input to that of \cmodel. It returns an answer of either \texttt{[True]} or \texttt{[False]}, indicating whether \model\ has answered the input question correctly.


To enhance the reliability of \vmodel, we use two methods: the Expanding Window (EW) method and the Multi-arm Bandits (MAB) with Upper Confidence Bound (UCB) \cite{slivkins2024introductionmultiarmedbandits}, as \emph{rejection filters} that filter potential false rejections of the \vmodel.


\underline{The EW method.}
The EW method starts by accepting the \vmodel\ LLM's predictions for $t$  TableQA instances, to calculate an initial accuracy of the LLM, denoted by $A(t)$: 
\begin{equation}
A(t) = \frac{\text{num\_correct}({M}_{Tw}, t)}{t}.
\end{equation}
Here, $\text{num\_correct}({M}_{Tw}, t)$ denotes the number of correct predictions made by the \vmodel\ LLM for the $t$ test instances. From the $(t+1)$-th test instance, the EW method rejects the LLM's \emph{unreliable} labels with a probability of $P(t) = 1 - A(t)$, and updates $A(t)$ to $A(t+1)$.

\underline{The MAB method.} The MAB method aims to balance exploration and exploitation by selecting actions that maximize expected rewards while considering uncertainty. We use an MAB of two arms representing either to accept or to reject an \emph{unreliable} label of the \vmodel\ LLM. We aim to select the arm that maximizes the TwAccuracy of \model, as guided by the equation below: 
\begin{equation}
\hat{\mu}_i(t) = \frac{\sum_{n=1}^{t-1} r_i(n) \cdot \mathbb{I}(a(n) = i)}{N_i(t -1)},
\end{equation}
where 
$\hat{\mu}_i(t)$ is the estimated mean reward of arm $i$ at time (i.e., test instance) $t$; $r_i(n)$ is the reward received when arm $i$ is selected at time $n$; $\mathbb{I}(a(n) = i)$ is the indicator function, which equals to 1 if arm $i$ is selected at time $n$, and 0 otherwise; and $N_i(t)$ is the number of times arm $i$ has been selected up to time $t$.

To balance exploitation (i.e., to follow the arm with a larger estimated mean reward $\hat{\mu}_i(t)$) and exploration (i.e., to try the other arm and accumulate more accurate mean reward estimates for the arm), we use the \emph{
Upper Confidence Bound} (UCB) algorithm: 
\begin{equation}
UCB_i(t) = \hat{\mu}_i(t) + c \cdot \sqrt{\frac{\ln t}{N_i(t)}},
\end{equation}
where $UCB_i(t)$ is the UCB for arm $i$ at time $t$; $c$ is the exploration parameter that controls the balance between exploration and exploitation; and $\ln$ is the natural logarithm function. We select the arm with a larger UCB value at time~$t$. 

\subsection{Fine-tuning for \cmodel{} and \vmodel{}}
\indent We adopt Low-Rank Adaptation (LoRA) \cite{hu2022lora}, to fine-tune our \cmodel\ and \vmodel{} with Llama3-8B-Instruct as the LLMs. 
The forward pass for both \cmodel\ and \vmodel{} LLMs is represented as:
\begin{align}
\resizebox{\columnwidth}{!}{$
    \mathbf{h} = \underbrace{\mathbf{W}_0 \mathbf{x}_{(T, Q, A)} + \Delta \mathbf{W} \mathbf{x}_{(T, Q, A)}}_{\text{gradient descent update}} =  \underbrace{\mathbf{W}_0 \mathbf{x}_{(T, Q, A)} + \mathbf{A} \mathbf{B} \mathbf{x}_{(T, Q, A)}}_{\text{LoRA update}}
    $},
\end{align}
where $\mathbf{W_0}$ is the initial parameter weights for the Transformers; $\mathbf{x}_{(T, Q, A)}$ is a tuple of input table, question, and model answer sets, including answers and their corresponding reasoning processes; and $\mathbf{\Delta W}$ is the gradient descent update for parameters which is decomposed into low-rank matrices $\mathbf{A}$ and $\mathbf{B}$:  $\mathbf{\Delta W} \in \mathbb{R}^{m \times d}$, $ \mathbf{A} \in \mathbb{R}^{m \times k}$, and $ \mathbf{B} \in \mathbb{R}^{k \times d}$. Matrices $\mathbf{A}$ and $\mathbf{B}$ have much fewer parameters compared with matrix $\mathbf{W}$, as $k \ll \min(m, d) $.

Since Llama3-8B-Instruct is used as the backbone model, the fine-tuning of \cmodel\ and \vmodel\ LLMs is a fully supervised classification task that is posed as an utterance prediction task of the label words. The aim of the LLMs is to maximize $P_\theta (x_t^{(i)} \,|\, x_1^{(i)}, x_2^{(i)}, \ldots, x_{t-1}^{(i)}) $, i.e., given previous tokens $(x_1^{(i)}, x_2^{(i)}, \ldots, x_{t-1}^{(i)})$, we maximize the conditional probability of $x_t^{(i)}$, where $\theta$ is the model parameter. Therefore, the loss functions for this fine-tuning task are:
\begin{align}
     \mathcal{L}_{{ans}} = - \sum_{i=1}^{V_{ans}} y_{m,i} \log(\hat{y}_{m,i});
     \mathcal{L}_{{twe}} = - \sum_{i=1}^{V_{twe}} y_{n,i} \log(\hat{y}_{n,i})
\end{align}
\vspace{-6mm}
\begin{align}
     \mathcal{L} & = - \log(\hat{y}_{m,y_m}) - \alpha \log(\hat{y}_{n,y_n}) = \mathcal{L}_{{ans}} + \alpha  \mathcal{L}_{{twe}}.
\end{align}
Here, $y_{m,i}$ and $y_{n,i}$ are the ground-truth labels for the $m$th and $n$th tokens at position $i$ in one-hot encoding; $\hat{y}_{m,i}$ and $\hat{y}_{n,i}$ are the model predicted probabilities for the $m$th and $n$th tokens at position $i$; $V_{ans}$ and $V_{twe}$ are the dictionary lengths of the \cmodel{} and  \vmodel{} LLMs, respectively; $\mathcal{L}_{{ans}}$ and $\mathcal{L}_{{twe}}$ are the loss functions for the two LLMs, respectively; and 
$\alpha$ is a coefficient used to control the influence of the \vmodel{} LLM on the result, with its default value being 1.

\begin{table}[ht!]
\centering
\setlength{\tabcolsep}{2pt}
\resizebox{\columnwidth}{!}{%
\begin{tabular}{lrrrrrr}
\toprule
\multicolumn{1}{c}{\multirow{2}{*}{\textbf{Dataset}}} &  \multicolumn{2}{c}{\textbf{\#  QA Pairs}}  & \multicolumn{2}{c}{\textbf{\# Numerical Questions}}   &  \textbf{Avg \# Tokens}  & \textbf{Avg \# Tokens} \\
\multicolumn{1}{c}{}    &  Training & Testing & Training & Testing & \textbf{per Table} &  \textbf{per Answer} \\     \midrule

\texttt{\wikidataset}    &   11,321    &   4,344     &  5,461  &  2,148 &  662.6 & 1.7                      \\ 
\midrule
\texttt{\datasetname} & 2,000 & 1,245 & 417 & 182 & 297.0 &5.1 \\
\midrule
\texttt{TabFact\_small} & 92,283 & 2,024 & 16,956 & 368 & 317.5 & 1.0 \\
\bottomrule
\end{tabular}
}
\caption{Dataset statistics. The number of QA pairs includes both numerical and non-numerical questions.} 
\label{tab:dataset-features}
\end{table}
\section{Experiments and Results}
Next, we present experimental results to verify the effectiveness of \model. We aim to answer the following questions: 
\textbf{(Q1)} How does \model\ compare with SOTA models and the latest LLMs, in terms of accuracy to process TableQA tasks?
\textbf{(Q2)} How effective is \model\ in tracking the trustworthiness of its answers? 
\textbf{(Q3)} How effective is \model\ in handling numerical problems?
\textbf{(Q4)} How much do the modules contribute to the overall accuracy of \model? 

\subsection{Experimental Setup}
\textbf{Datasets.} We use a public benchmark dataset named \texttt{WikiTableQuestions} (denoted as \textbf{\texttt{\wikidataset}})~\cite{pasupat-liang-2015-compositional} and 
a dataset named \textbf{\texttt{\datasetname}} which we adapted from the \texttt{FeTaQA} dataset~\cite{nan2022fetaqa}. Besides, we use a third dataset named \textbf{\texttt{TabFact}}~\cite{ChenWCZWLZW20} to showcase the general applicability of \model. 

\begin{figure}[ht!]
     \centering
     \includegraphics[width = 0.9\linewidth]{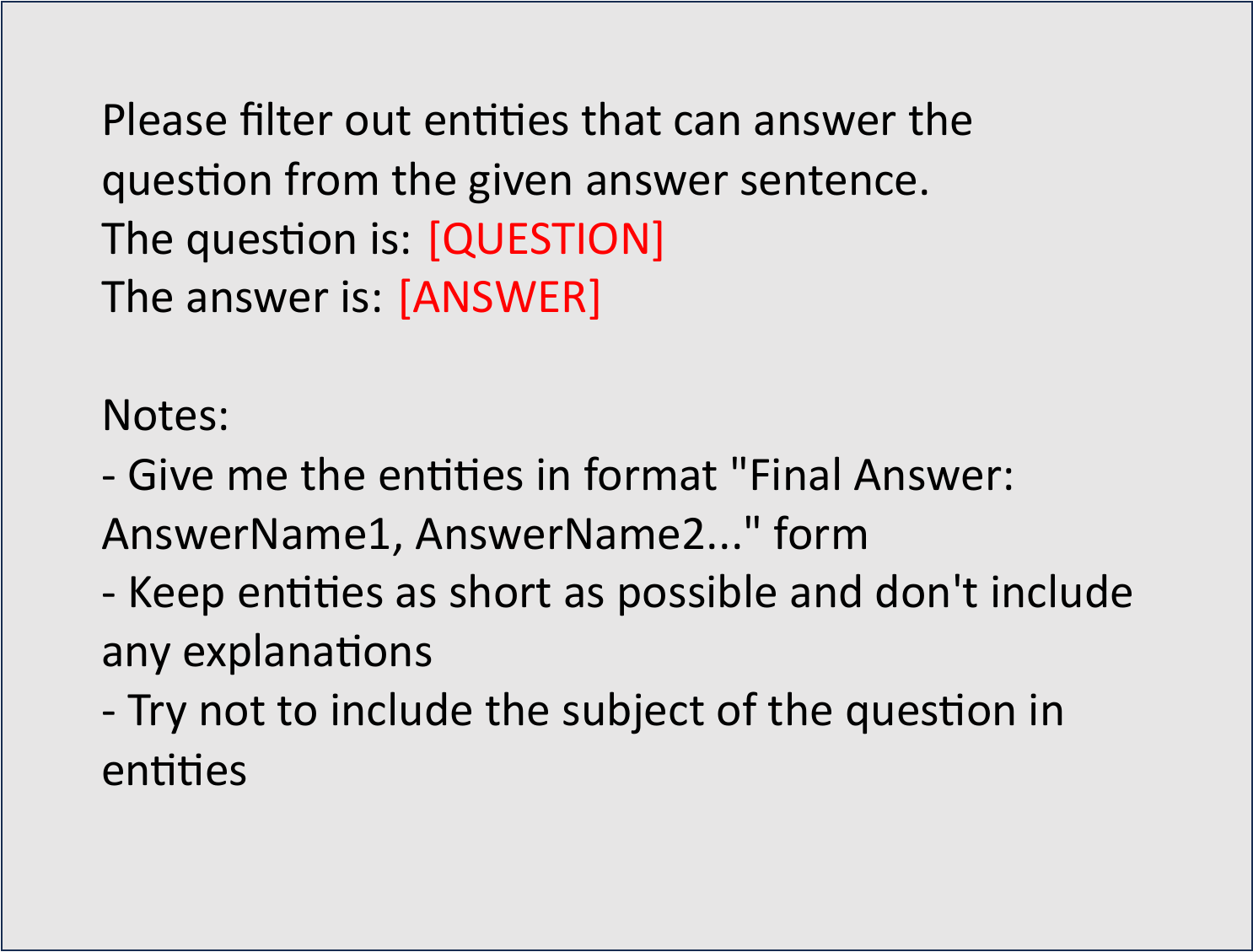}
     \caption{Prompt for GPT-4o to extract key entities from the \texttt{FeTaQA} dataset.}
     \label{fig:filter}
\end{figure}

\texttt{FeTaQA} is also a TableQA dataset. Its answers contain descriptive content that may not be directly relevant to the questions. We remove such noisy information from the answers and only retain the entities that answer the questions, to form  the \texttt{\datasetname} dataset. This makes automatic evaluation of model accuracy on the dataset easier and more precise (instead of using fuzzy metrics such as the ROUGE score). 
We use GPT-4o to extract key entities from the answers that are relevant to the questions, which recorded an accuracy of 67.7\%. We manually check the extracted entities and correct them if necessary to ensure the data quality. The prompt used is shown in Figure~\ref{fig:filter}.

The original \texttt{FeTaQA} dataset contains questions on Wikipedia tables where the ground-truth answers are in free-form text, with an average of 18.9 tokens per answer. After our clean-up process, the ground-truth answers in our \texttt{\datasetname} dataset consist of multiple entities that directly answer the questions, with an average of 5.1 tokens per answer.
Table~\ref{tab:answer_comparison} shows an example of a question and its answer in  \texttt{FeTaQA} and \texttt{\datasetname}.


\begin{table}[h!]
\centering
\setlength{\tabcolsep}{2pt}
\resizebox{0.47\textwidth}{!}{
\begin{tabular}{ll}
\toprule
\textbf{Dataset} & \textbf{Answer} \\
\midrule
\texttt{FeTaQA} & \begin{tabular}[c]{@{}l@{}}In 2019, Shagun Sharma played in the roles as Pernia in \\ \underline{Laal Ishq}, \underline{Vikram Betaal Ki Rahasya Gatha} as \\ Rukmani/Kashi and \underline{Shaadi Ke Siyape} as Dua.\end{tabular} \\
\midrule
\texttt{\datasetname} & \begin{tabular}[c]{@{}l@{}}Laal Ishq, Vikram Betaal Ki Rahasya Gatha,\\ Shaadi Ke Siyape\end{tabular} \\
\bottomrule
\end{tabular}
}
\caption{Comparison of answers in \texttt{FeTaQA} and our \texttt{\datasetname} datasets, for the question ``What TV shows was Shagun Sharma seen in 2019?".} 
\label{tab:answer_comparison}
\end{table}

Table~\ref{tab:dataset-features} summarizes the dataset statistics. 
\texttt{\wikidataset} is a larger dataset with longer tables, while \texttt{\datasetname} has longer answers on average, making it a more challenging dataset (because its questions may contain multiple sub-questions). \texttt{TabFact\_small} is a subset of the \texttt{TabFact} dataset for table-based fact verification, where the task is to determine whether each statement is entailed or refuted.

Table~\ref{tab:token_dist} shows the distribution of 
the table length (i.e., number of tokens per table) of the \texttt{\wikidataset}, 
\texttt{\datasetname}, and \texttt{TabFact\_small} datasets. The tables in \texttt{\wikidataset} are larger compared with the other two datasets, about $31.4\%$ of which have over 500 tokens, while 83.2\% of the tables in \texttt{\datasetname} have fewer than 500 tokens, and 82.7\% in \texttt{TabFact\_small}.

\begin{table*}[ht!]
\centering
\setlength{\tabcolsep}{3pt}
\begin{tabular}{ccccc}
\hline
\multicolumn{1}{c}{\multirow{2}{*}{\textbf{Dataset}}} & \multicolumn{4}{c}{\textbf{\# tokens per table (\%)}} \\
& Below $500$ & $500-1000$  & $1000-2000$ & Above $2000$\\ 
 \hline
\texttt{\wikidataset} & $68.6$ & $16.8$ & $8.9$ & $5.7$ \\
\hline
\texttt{\datasetname} & $83.2$ & $13.7$ & $2.6$ & $0.5$ \\
\hline
\texttt{TabFact\_small} & $82.7$ & $15.4$ & $1.8$ & $0.1$ 
\\ \hline
\end{tabular}
\caption{Distribution of tables with different \# tokens of the  \texttt{\wikidataset}, \texttt{\datasetname}  and \texttt{TabFact\_small} datasets.}
\label{tab:token_dist}
\end{table*}

The numerical questions are filtered and counted by keyword matches. Details are shown below:

\textbf{Numerical Question Filtering.} We collect frequently used keywords for the numerical questions, including ``how many", ``number", ``the most", ``difference", ``count", ``highest'', ``average", ``at least'', ``rank", ``lowest", ``percentage", ``sum", ``compare", and ``frequency" to extract potential numerical questions. The keyword counts are shown in Table~\ref{tab:keywords_cnt}.

\begin{table*}[ht!]
\centering
\setlength{\tabcolsep}{3pt}
\resizebox{\textwidth}{!}{%
\begin{tabular}{ccccccccccccccc}
\hline
\multicolumn{1}{c}{\multirow{2}{*}{\textbf{Dataset}}} & \multicolumn{14}{c}{\textbf{Numerical question related keywords}} \\
 & how many & number & the most & difference & count & highest & average & at least & rank & lowest & percentage & sum & compare & frequency \\ 
 \hline
\texttt{\wikidataset} & 4,097 & 1,912 & 814 & 410 & 399 & 215 & 153 & 149 & 145 & 81 & 73 & 48 & 16 & 4  \\
\hline
\texttt{\datasetname} & 369 & 43 & 16 & 4 & 12 & 18 & 10 & 3 & 13 & 3 & 51 & 43 & 26 & - \\ \hline
\texttt{TabFact\_small} & 9 & 4998 & 2195 & 400 & 1613 & 2898 & 925 & 254 & 2119 & 1464 & 238 & 329 & 84 & 298 \\
\hline
\end{tabular}
}
\caption{Numerical question related keyword counts of the  \texttt{\wikidataset}, \texttt{\datasetname}, and \texttt{TabFact\_small} datasets.}
\label{tab:keywords_cnt}
\end{table*}

Each keyword may suggest single- or multi-step calculations, while a question with multiple keyword matches typically involve multi-step calculations. 

\textbf{Examples of Multi-hop Numerical Problems.} Below, we show three typical types of multi-hop numerical questions found in the datasets, the answers of which cannot be obtained through a one-step calculation. 
\begin{enumerate}
\item Ordering comparison after some calculation:\\
\textit{Which country had the most riders that placed in the top 20 of the 1971 trans-ama final standings?}

\item  Difference comparison after some calculation:\\
\textit{What's the difference between the highest score in a game and the lowest score in a game?}

\item  Average (i.e., sum and division) calculation: \\
\textit{What is the average number of draws?}

\end{enumerate}
Among the  \texttt{\wikidataset} and \texttt{\datasetname} datasets, there are 814 and 16 Type-1 questions, 410 and 4 Type-2 questions, and 153 and 10 Type-3 questions, respectively. For \texttt{TabFact\_small} (including training set of \texttt{TabFact}), there are 2195 Type-1 questions, 34 Type-2 questions, and 925 Type-3 questions, respectively.

\begin{table}[ht!]
\centering
\setlength{\tabcolsep}{2pt}
\resizebox{\columnwidth}{!}{%
\begin{tabular}{llrrrr}
\toprule
\multicolumn{2}{c}{\multirow{2}{*}{\textbf{Model}}} & \multicolumn{2}{c}{\textbf{Accuracy (\%)}} & \multicolumn{2}{c}{\textbf{TwAccuracy (\%)}}  \\
&\multicolumn{1}{c}{} & \texttt{\wikidataset} & \texttt{\datasetname} & \texttt{\wikidataset} & \texttt{\datasetname} \\
\midrule
{\multirow{2}{*}{Fine-tuned PLMs}} & TAPEX-Large      &  $57.5$  & $19.8$ & ~~{--} & ~~{--} \\
    & OmniTab-Large      &  $63.3$  & $25.2$ & ~~{--} & ~~{--} \\
 \midrule
{\multirow{3}{*}{Zero-shot LLMs}} & GPT-3.5 Turbo & $50.9$ &  $38.1$ & ~~{--} & ~~{--}\\
& GPT-4o & $58.1$ &  \underline{$41.7$} & ~~{--} & ~~{--} \\
& Mix-SC & \underline{$72.5$} & $41.6$ & ~~{--} & ~~{--} \\
\midrule
{\multirow{3}{*}{Few-shot LLMs}} & Binder  & $64.6$ & ~~{--} & ~~{--} & ~~{--} \\
& DATER  & $65.9$ & ~~{--} & ~~{--} & ~~{--} \\
& Chain-of-Table  & $67.3$ & ~~{--} & ~~{--} & ~~{--} \\
\midrule
{\multirow{3}{*}{Ours}} & \model-EW & $\mathbf{76.6}$  & $\mathbf{44.1}$  & $77.6$ & $\mathbf{53.8}$\\
&\model{} & $\mathbf{76.6}$ & $\mathbf{44.1}$ & $\mathbf{77.9}$ & $53.7$\\
& $\Delta$ & $+5.7$ & $+5.8$ & & \\
\bottomrule
\end{tabular}
}
\caption{TableQA performance results on \texttt{\wikidataset} and \texttt{\datasetname} datasets. 
Best results are in \textbf{boldface}, while second best results are \underline{underlined}; $\Delta$ (\%) denotes the performance gain of \model\  comparing with the best baseline results.}
\label{tab:main_results}
\end{table}

\textbf{Model input data preparation.} Following a common practice of the literature~\cite{yin-etal-2020-tabert, liu-etal-2024-rethinking, wang2024chainoftable}, tables are flattened and converted into a sequence to form part of the prompts to the LLMs.  Table cells are separated by `$\vert$' characters, while rows are separated by line breaks. Questions in natural language are added directly into the prompts, and answers (which are also in natural language) are only used for training the \cmodel\ and the \vmodel\ modules. Details of the ground-truth construction for fine-tuning \cmodel\ and \vmodel{} are shown below:

\textbf{Ground-truth Label Construction for Training \cmodel\ and \vmodel.} For \cmodel\ and \vmodel{} training, we collect the training output of both model branches (i.e., SOTA TableQA model and our \nmodel). Given a training instance, if only one model branch yields the correct result, its ground-truth label for \cmodel\ is that branch and the ground-truth label for \vmodel\ is a \texttt{[True]} label (detailed later). If both branches return the correct answer, we randomly label one of the branches as the ground-truth label for the \cmodel\ based on the overall accuracy of the two branches, while the ground-truth label for \vmodel\ is again the  \texttt{[True]} label.
If both branches return a wrong answer, we discard this training instance when training \cmodel, while the ground-truth label of this instance for \vmodel{} is \texttt{[False]}.

\begin{figure}[ht!]
     \centering
     \includegraphics[width = 1 \linewidth]{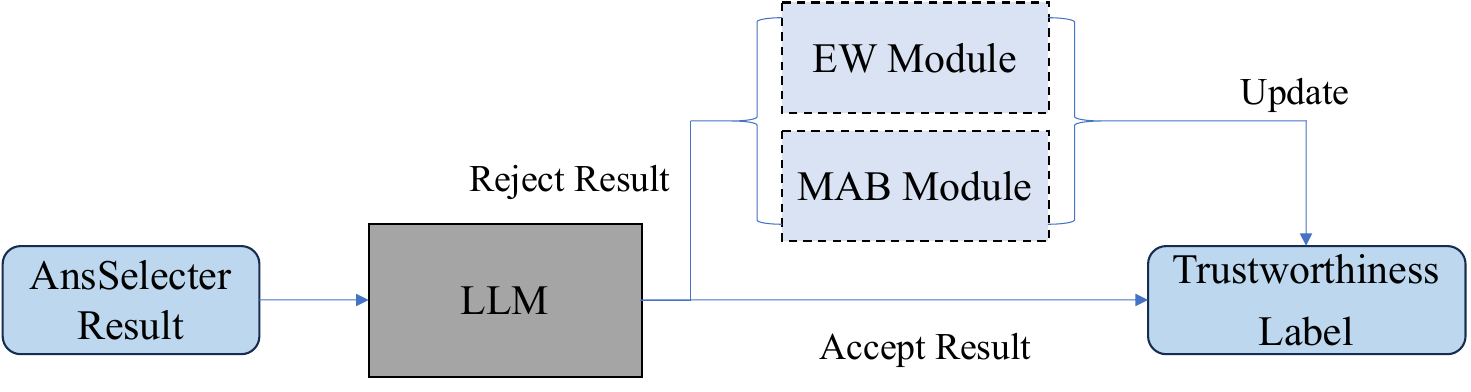}
     \caption{Structure of the \vmodel\ module.}
     \label{fig:tw_eval_inter}
\end{figure}

For the \cmodel\ and \vmodel\ LLMs, they take the question, the table titles and headers but not the contents as input, together with the answers and reasoning process, such that they can focus on analyzing the generated answers instead of attempting to answer the questions again. As mentioned earlier, the prompts for the LLMs of \model, \cmodel, and \vmodel{} can be found in section~\nameref{sec:ans_gen} above. 

\textbf{Competitors.} We compare our models \model-EW (which uses the EW method for \vmodel) and \model\ with three categories of baseline models including SOTA in each category: (1)~Fine-tuned PLM-based models: TAPEX-Large~\cite{liu2022tapex} and OmniTab-Large~\cite{jiang-etal-2022-omnitab} (SOTA); (2)~Zero-shot LLM-based models: GPT-3.5-Turbo~\cite{openai2024gpt35}, GPT-4o~\cite{openai2024gpt4o}, and Mix-SC~\cite{liu-etal-2024-rethinking} (SOTA); (3)~Few-shot LLM-based models: Binder~\cite{cheng2023binding}, DATER~\cite{ye2023large}, and Chain-of-Table~\cite{wang2024chainoftable} (SOTA). 
For Binder, DATER, and Chain-of-Tables, we use the best results reported in their papers. For the other models, we rerun the experiments on the three datasets (with fine-tuning if applicable).

\textbf{Implementation details.} We use Mix-SC for the SOTA branch of \model\ and 
GPT-3.5 Turbo as the backbone model of \nmodel. We fine-tune the \cmodel\ and \vmodel\ LLMs with the AdamW optimizer~\cite{loshchilov2019decoupledweightdecayregularization} using a learning rate of 0.0002 and a weight decay 0.001. The maximum number of input tokens is 5,000, and the maximum number of epochs is 20. 

\textbf{Model Fine-tuning Hyper-parameters and Performance.} Table~\ref{tab:peft} summarizes the hyper-parameter values used for training  \cmodel\ and (the LLM of) \vmodel. We share the parameter values between both modules for consistency and reproducibility. 

\begin{table}[ht!]
\centering 
\begin{tabular}{lr}
\toprule
\textbf{Hyper-parameter}                            & \textbf{Value}  \\
\toprule
lora\_alpha          &  64 \\
r       & 16 \\
optimizer                                & paged\_adamw\_32bit          \\
learning rate                & 2e-4       \\
weight decay                   & 0.001       \\
\(max.\) gradient norm                 & 0.3      \\
warm-up ratio              & 0.03       \\
\toprule
\end{tabular}
\caption{Module fine-tuning hyper-parameters for \cmodel\ and \vmodel.}
\label{tab:peft}
\end{table}

\textbf{Evaluation metrics.} We report both the exact-match answer 
 \textbf{Accuracy} 
 for each model, as well as the accuracy (\textbf{TwAccuracy}) of the trustworthiness label generated by our \vmodel\ module.
TwAccuracy is calculated as the number of times when the  trustworthiness label is correctly predicted, divided by the total number of test instances.

All experiments are run with two NVIDIA A100 80 GB GPUs on a cloud GPU server. 

\subsection{Results} 

\textbf{Overall results (Q1).} Table~\ref{tab:main_results} reports the overall performance of the models. Our \model\ models outperform all competitors on both datasets, improving TableQA accuracy by 5.7\% and 5.8\%, respectively. This confirms the effectiveness of our dual-branch model structure to exploit both a SOTA TableQA model and our \nmodel\ to yield accurate answers for more questions. We run chi-squared tests comparing \model\ with SOTA (Mix-SC on \texttt{\wikidataset} and GPT-4o on \texttt{\datasetname}), yielding p-val of $0.002$ and $0.24$, respectively. This confirms that the result on \texttt{\wikidataset} is statistically significant, while the larger p-val on \texttt{\datasetname} is due to the smaller test set size. 
The accuracy of \model-EW and \model\ are the same because they share the answer generation modules and differ only in the \vmodel\ module (detailed next). 

Among the baseline models, Mix-SC has the best overall results, for its self-consistency-based method to choose the most likely answer from multiple answers.  
On \texttt{\datasetname}, GPT-4o is slightly better than Mix-SC, because Mix-SC uses a less advanced backbone, GPT-3.5 Turbo, while we have used its default prompts which might not be optimized for \texttt{\datasetname}.  

The PLM-based models are uncompetitive. Their smaller model sizes limited their semantic understanding capability. 

The few-shot LLM-based models Binder, DATER, and Chain-of-Tables do \emph{not} run directly on \texttt{\datasetname}. Since they are outperformed by Mix-SC, we did not adapt their implementation for comparison on \texttt{\datasetname}.

\textbf{Additional Results on TabFact\_small.} To show the general applicability of \model, we conduct experiments on the \texttt{TabFact\_small} dataset. The experimental results are shown in Table~\ref{tab:tabfact}. Here, both our \nmodel and \model\ use GPT-4o mini as the backbone model, because GPT-4o mini's Python code generation performance on this dataset is significantly better than that of GPT-3.5 Turbo.  

\begin{table}[ht!]
\centering
\setlength{\tabcolsep}{2pt}
\resizebox{\columnwidth}{!}{%
\begin{tabular}{llcc}
\toprule
\multicolumn{2}{c}{\textbf{Model}} & \textbf{Accuracy (\%)} & \textbf{TwAccuracy (\%)} \\
\midrule
{\multirow{2}{*}{Fine-tuned PLMs}} & TAPEX-Large  &  $75.3$ & ~~{--} \\
& OmniTab-Large  &  $85.2$  & ~~{--} \\
 \midrule
{\multirow{2}{*}{Zero-shot LLMs}} & GPT-3.5 Turbo & $82.5$ & ~~{--} \\
& GPT-4o mini & \underline{$88.6$} & ~~{--} \\
\midrule
{\multirow{2}{*}{Few-shot LLMs}} & DATER  & $85.6$ & ~~{--} \\
& Chain-of-Table  & $86.6$ & ~~{--} \\
\midrule
{\multirow{2}{*}{Ours}} & \nmodel & $89.0$ & ~~{--} \\
&\model{} & $\mathbf{90.4}$ & $\mathbf{90.0}$\\
& $\Delta$ & $+2.0$ & ~~{--} \\
\bottomrule
\end{tabular}
}
\caption{TableQA performance results on the \texttt{TabFact\_small} dataset. 
Best results are in \textbf{boldface}, while second best results are \underline{underlined}; $\Delta$ (\%) denotes the performance gain of \model\  comparing with the best baseline results.}
\label{tab:tabfact}
\end{table}

\begin{figure}[ht!]
     \centering
     \includegraphics[width = 1\linewidth]{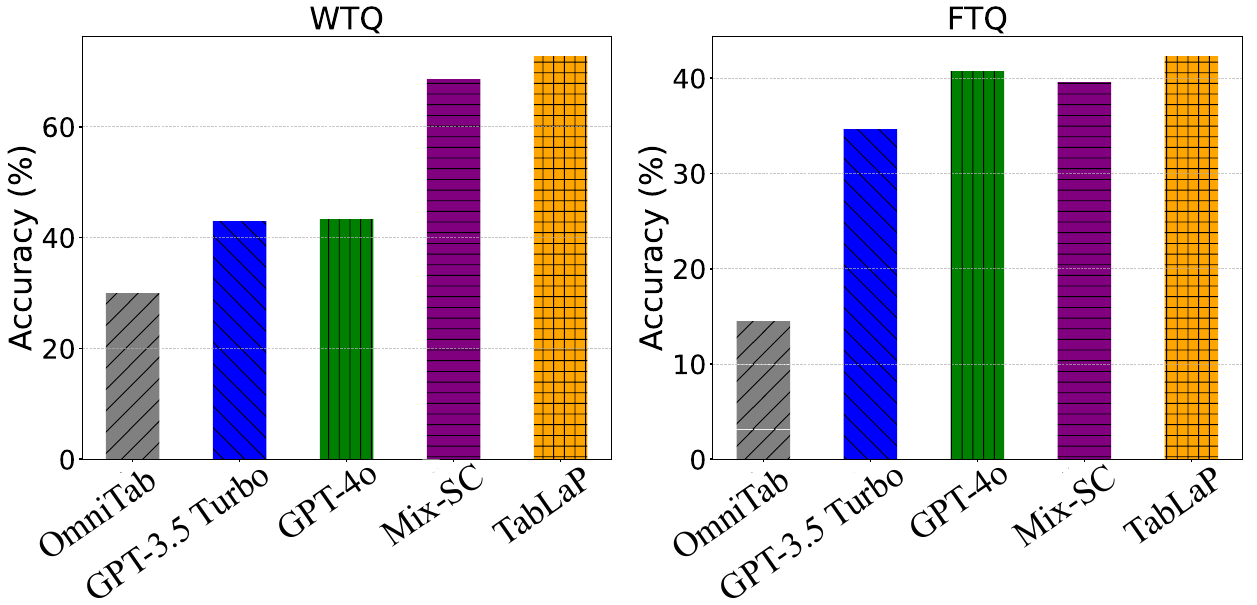}
     \caption{Model accuracy on numerical questions.}
     \label{fig:math_problem}
\end{figure}

\textbf{Effectiveness in tracking answer trustworthiness (Q2).} Table~\ref{tab:main_results} also shows   the accuracy (i.e., TwAccuracy) of \model\ to report the trustworthiness of its answers. On \texttt{\wikidataset}, TwAccuracy of the \model\ models is close to 80\%, meaning that users can follow the trustworthiness labels to either consume or reject our models' answers, without any regret for four out of five questions asked to \model\ on average. 

On \texttt{\datasetname}, TwAccuracy of the \model\ models is not as high, because the dataset is more difficult, such that it is also challenging to predict the trustworthiness of the model answers. Note that TwAccuracy of \model\ is now much higher than the answer accuracy (53.7\% vs 44.1\%), meaning that it is more beneficial to follow our trustworthiness labels than blindly trusting the answers.  

None of the baseline models offer any trustworthiness labels and hence they do not have TwAccuracy results. If users simply accept the answers generated by these models, the TwAccuracy of these models will be the same as their answer accuracy. Let the \emph{user regret ratio} of a model be $1 - TwAccuracy$. Then, the lowest regret ratios of the baselines are $1 - 72.5\% = 27.5\%$ and $1 - 41.7\% = 58.3\%$ on the two datasets, respectively, while those of \model\ are $1 - 77.9\% = 22.1\%$ and $1 - 53.7\% = 46.3\%$, which are 19.6\% and 20.6\% lower, respectively.

\textbf{Performance on numerical questions (Q3).} Figure~\ref{fig:math_problem} reports accuracy on numerical questions. We show the results of the SOTA PLM-based model (OmniTab) and zero-shot LLMs (GPT-4o and Mix-SC). We also include  GPT-3.5 Turbo since it is the backbone of our \nmodel.


\model\ has the highest accuracy on both datasets, now outperforming the best baselines by 6.3\% (72.8\% vs. 68.5\% of Mix-SC) and 6.8\% (42.3\% vs. 39.6\% of GPT-4o) on the two datasets, respectively. 
This result confirms that exploiting the reasoning and planning capabilities of the backbone LLM to process numerical questions is more effective  than prompting the LLM to generate the results directly. 


\begin{table}[ht!]
\centering
\setlength{\tabcolsep}{2pt}
\begin{tabular}{llrr}
\toprule
\multicolumn{1}{c}{\multirow{2}{*}{\textbf{Model}}} & \multicolumn{2}{c}{\textbf{Accuracy (\%)}}  \\
  & \texttt{\wikidataset} & \texttt{\datasetname} \\
\midrule
\nmodel & $64.7$ & $42.6$  \\
\model{}-w/o-\nmodel & $62.2$ & $43.5$  \\
\model{}-w/o-\cmodel & $68.5$ & $42.2$  \\
\model{} & $\mathbf{76.6}$ & $\mathbf{44.1}$  \\
\bottomrule
\end{tabular}
\caption{Ablation study results.}
\label{tab:ablation}
\end{table}

\textbf{Ablation study (Q4).} 
We conduct an ablation study with four model variants: (1)~\nmodel; 
(2)~\model-w/o-\nmodel, where \nmodel\ is replaced with GPT-3.5 Turbo; (3)~\model-w/o-\cmodel, where \cmodel\ is replaced with a random selection of the answers from the two branches of \model; and (4)~\model. 

As Table~\ref{tab:ablation} shows, using just \nmodel\ causes substantial accuracy drops, as there are many non-numerical questions in the datasets which \nmodel\ is not designed for. Meanwhile, removing either \nmodel\ or \cmodel\ from \model\ also leads to lower accuracy. This confirms the effectiveness of both modules in contributing to the overall accuracy of \model. We also replace our \nmodel\ and \vmodel\ with two out-of-box LLMs, Llama3-8B-Instruct and GPT4o-mini, which result in lower Accuracy as well (see in Table~\ref{tab:replace_selector}), confirming the need for the fine-tuned modules.

\begin{table}[ht!]
\centering
\setlength{\tabcolsep}{2pt}
\resizebox{\columnwidth}{!}{%
\begin{tabular}{lrrrr}
\toprule
\multicolumn{1}{c}{\multirow{2}{*}{\textbf{Model}}} & \multicolumn{2}{c}{\textbf{Accuracy (\%)}} & \multicolumn{2}{c}{\textbf{TwAccuracy (\%)}}  \\
& \texttt{\wikidataset} & \texttt{\datasetname} & \texttt{\wikidataset} & \texttt{\datasetname} \\
\midrule
Llama3-8B Instruct w/o fine-tuned & $66.4$ & $40.7$ & $59.7$ & $46.1$ \\
GPT-4o mini & $68.6$ & $43.1$ & $75.7$ & $47.4$ \\
\model\ (ours) & $\mathbf{76.6}$ & $\mathbf{44.1}$ & $\mathbf{77.9}$ & $\mathbf{53.7}$ \\
\bottomrule
\end{tabular}
}
\caption{TableQA performance results on \texttt{\wikidataset} and \texttt{\datasetname} datasets using different LLMs as \cmodel\ and \vmodel.}
\label{tab:replace_selector}
\end{table}

\begin{figure}[htp]
    \centering
    \begin{subfigure}[a]{\linewidth}
        \centering
        \includegraphics[width=0.8\linewidth]{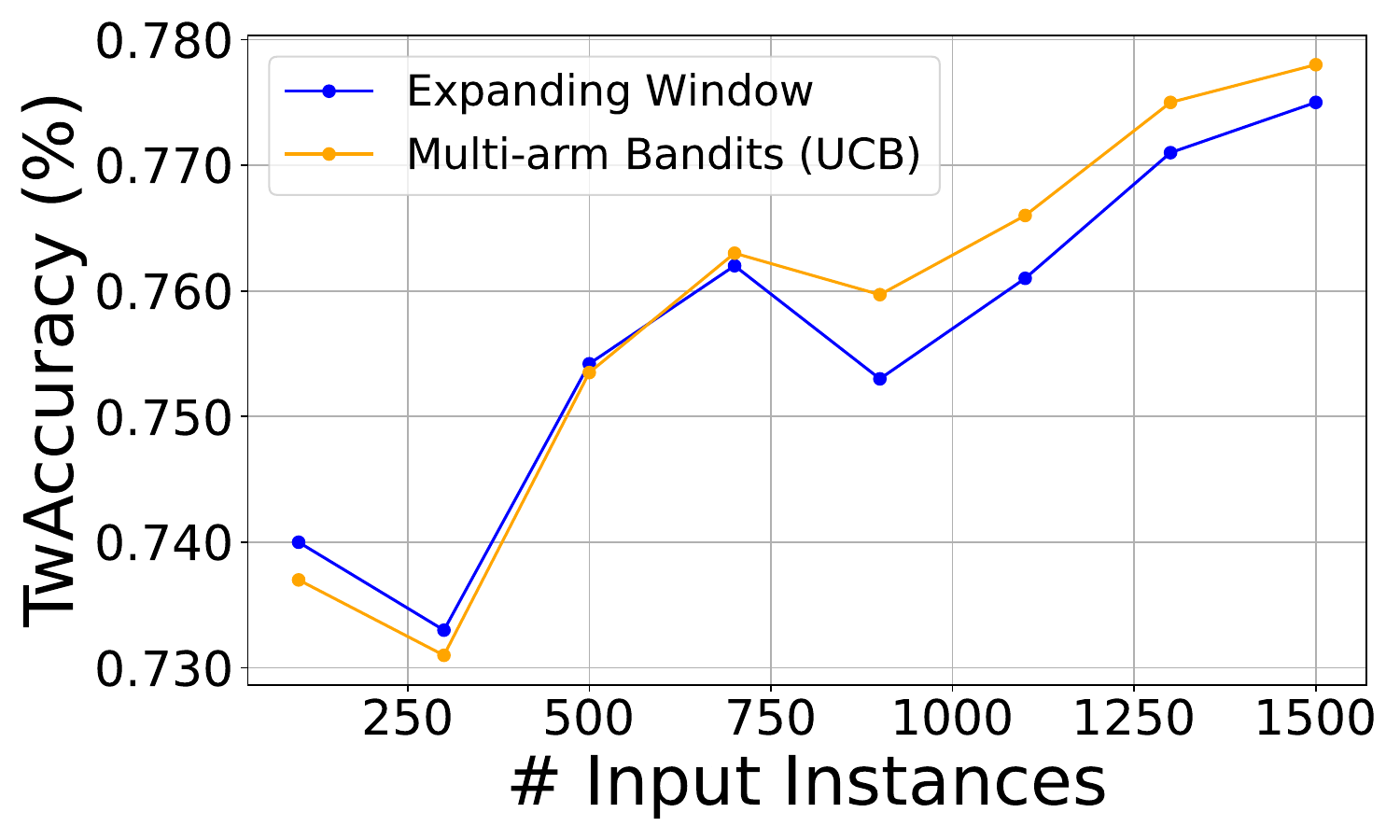}
        \caption{\texttt{\wikidataset}}
        \label{fig:tw_acc_a}
    \end{subfigure}
    \begin{subfigure}[a]{\linewidth}
        \centering
        \includegraphics[width=0.8\linewidth]{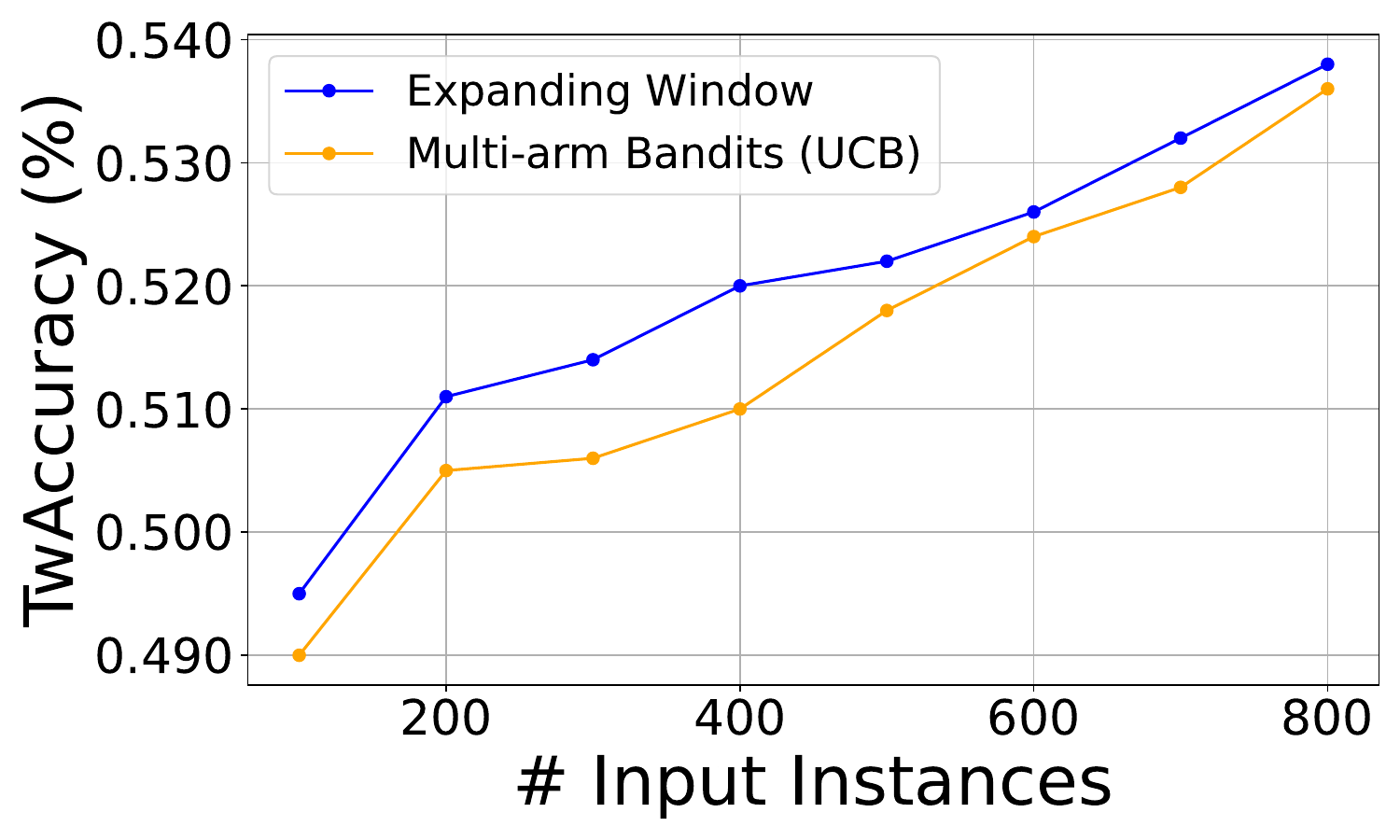}
        \caption{\texttt{\datasetname}}
        \label{fig:tw_acc_b}
    \end{subfigure}
    \caption{\vmodel's TwAccuracy with the EW and MAB methods over different numbers of test instances.}
    \label{fig:tw_acc}
\end{figure}

We then study the effectiveness of \vmodel\ with EW and MAB. As shown in Table~\ref{tab:main_results}, both \model-EW and \model\ (with MAB) share similar TwAccuracy. While \model-EW is more accurate on \texttt{\datasetname}, \model\ outperforms on \texttt{\wikidataset}. This is because \model-EW can estimate the accuracy of the \vmodel\ LLM quickly by the simple design of EW, which better suits the smaller test set of \texttt{\datasetname}, while MAB takes more test instances to learn an (more precise) estimation of the accuracy of the \vmodel\ LLM -- refer to the results below:

\textbf{Effectiveness of EW and MAB.} Figure~\ref{fig:tw_acc} shows the performance of \vmodel\ using the EW and the MAB methods, respectively. The EW method helps achieve high TwAccuracy with fewer test instances, while the TwAccuracy of MAB grows slower but may reach even higher values when more test instances are seen. This is consistent with the design goal of the MAB, i.e., to obtain overall better results through a progressive exploration and exploitation process. This result also suggests an optimization opportunity: to start with EW  while training MAB in the background; once MAB has gained higher accuracy, we switch to it.  

\begin{figure}[ht!]
    \centering
    \begin{subfigure}[a]{\linewidth}
        \centering
        \includegraphics[width=\linewidth]{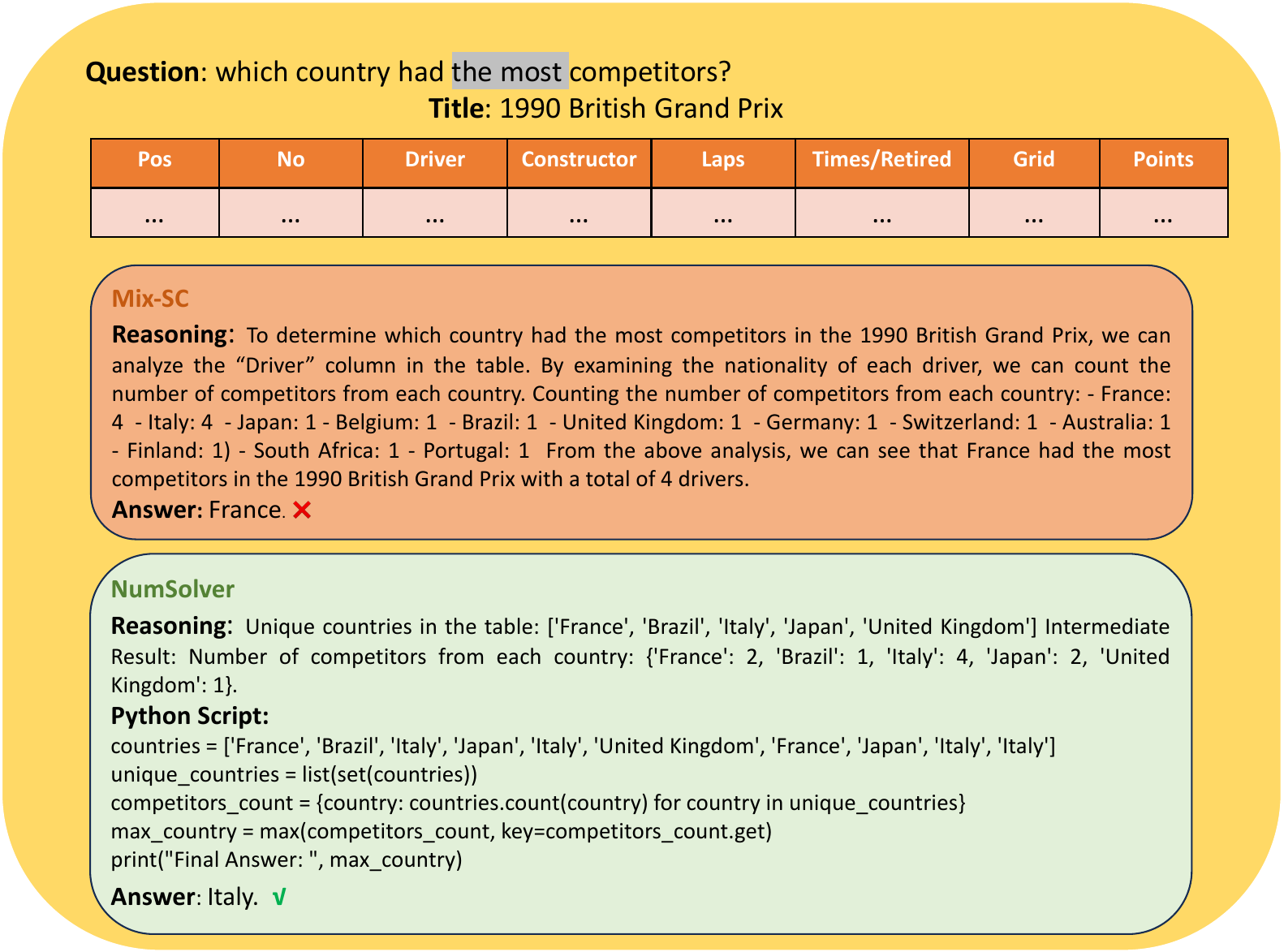}
        \caption{A question to compare order after summation.}
        \label{fig:math_sub2}
    \end{subfigure}
    \begin{subfigure}[b]{\linewidth}
        \centering
        \includegraphics[width=\linewidth]{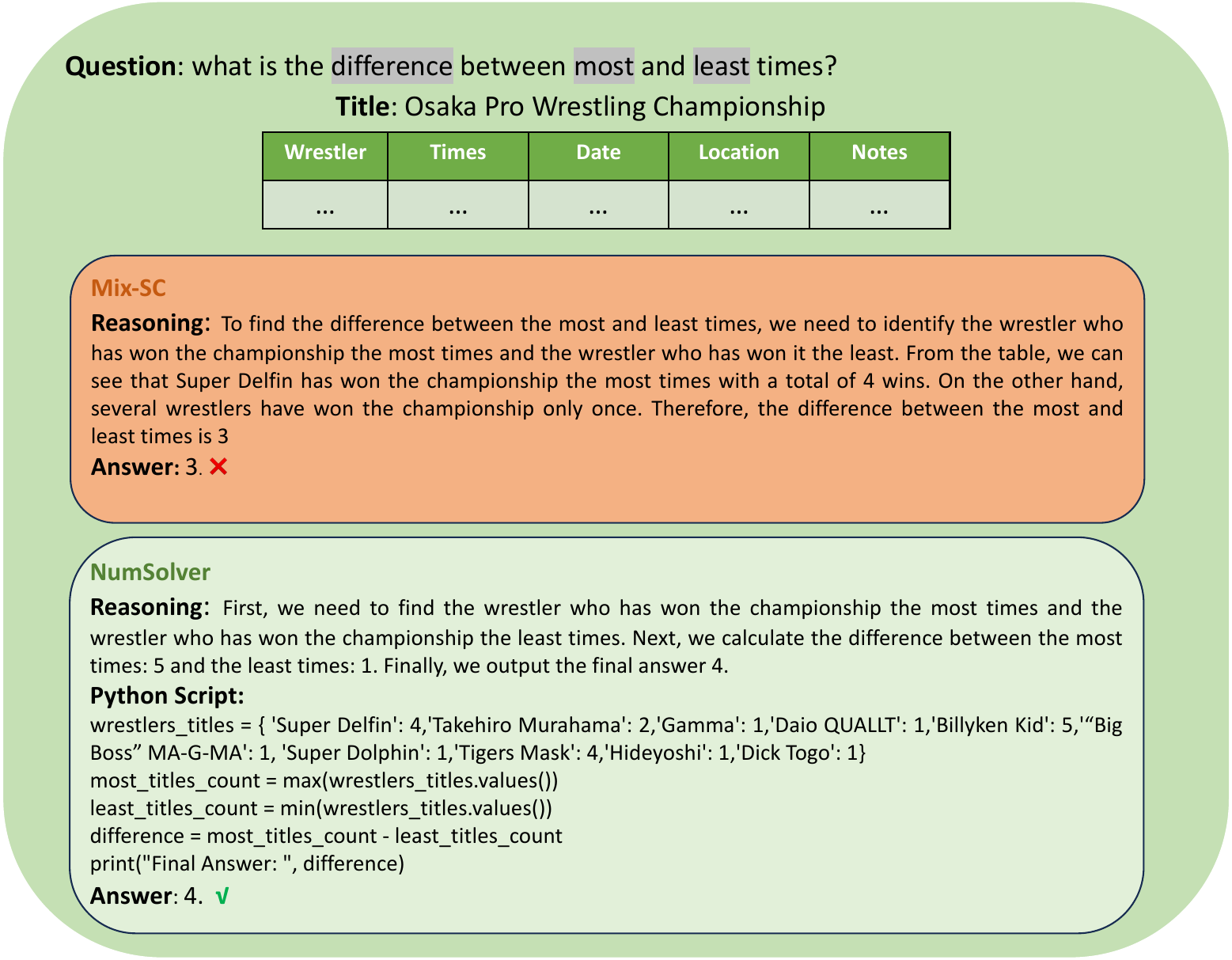}
        \caption{A question to calculate difference after summation.}
        \label{fig:math_sub3}
    \end{subfigure}
    \begin{subfigure}[c]{\linewidth}
        \centering
    \includegraphics[width=\linewidth]{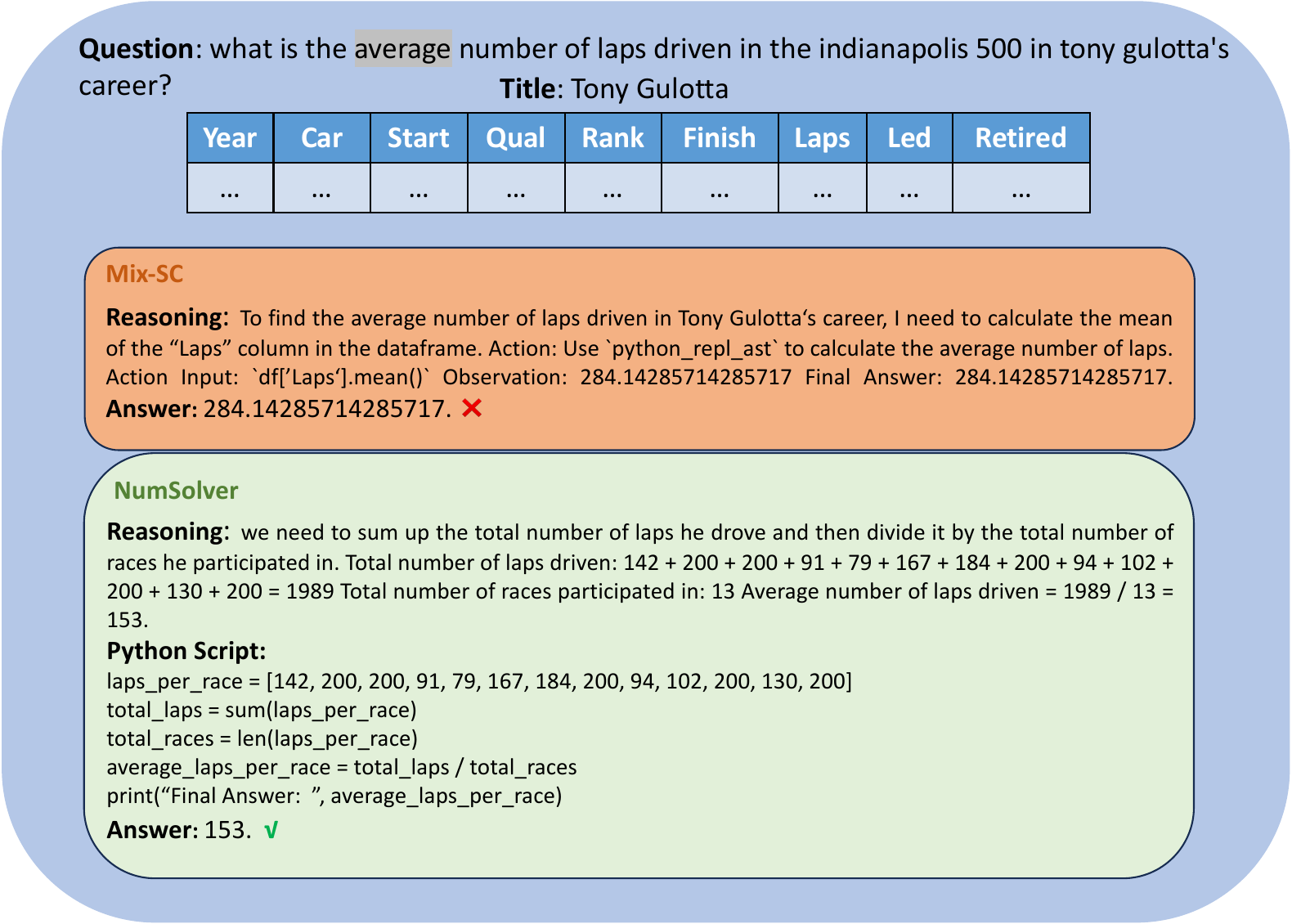}
        \caption{A question to calculate average (i.e., summation and division).}
        \label{fig:math_sub1}
    \end{subfigure}
    \caption{Three examples of multi-hop numerical questions and answers generated by the SOTA model Mix-SC and by our \nmodel.}
    \label{fig:case_study}
\end{figure}

\textbf{Case study.} We also include three typical multi-hop numerical questions that fail the SOTA model Mix-SC, while \model\ successfully answers those questions in Figure~\ref{fig:case_study}. 

\section{Conclusion}
 We proposed \model{}, an accurate and regret-aware multi-LLM-based model for the TableQA task. \model\ uses an LLM as a planner to generate calculation plans for numerical questions, exploiting the reasoning capability of LLMs while avoiding their limitations in carrying out the actual calculations. \model\ comes with a module based on multi-arm bandit to quantify the trustworthiness of the answers generated by the model, enabling users to consume the answers in a regret-aware manner for the first time. We verified the effectiveness of \model\ on a public benchmark dataset  
 \texttt{WikiTableQuestions} and an adapted dataset \texttt{\datasetname}. The results show that \model\ outperforms SOTA TableQA models in accuracy by 5.7\% and 5.8\% on the two datasets, respectively. Meanwhile, the answer trustworthiness labels generated by \model\ help reduce the user regret ratio on consuming the model generated answers by 19.6\% and 20.6\% on the two datasets, respectively, compared with always trusting the model generated answers.   

 \section{Acknowledgments}
This work is in part supported by the Australian Research Council (ARC) via Discovery Projects DP230101534 and DP240101006. Jianzhong Qi is supported by ARC Future Fellowship FT240100170. Junhao Gan is in part supported by ARC Discovery Project DP230102908.

\newpage

\bibliography{aaai25}

\end{document}